\documentclass{article}
\usepackage[a4paper,margin=3cm]{geometry}

\usepackage{amsmath,amsfonts,bm}

\def\eqref#1{equation~\ref{#1}}

\def\1{\bm{1}}

\DeclareMathAlphabet{\mathsfit}{\encodingdefault}{\sfdefault}{m}{sl}
\SetMathAlphabet{\mathsfit}{bold}{\encodingdefault}{\sfdefault}{bx}{n}

\usepackage{amssymb}
\usepackage{graphicx}
\usepackage{booktabs,tabularx,array}
\usepackage{xcolor}
\usepackage{hyperref}
\usepackage{url}
\usepackage[numbers,sort&compress]{natbib}
\usepackage{doi}
\usepackage{authblk}
\hypersetup{
  pdftitle={CAFE+FNO: Fourier Kernel Generation via Multiplicative Feature Composition},
  pdfauthor={Hyungjoon Juen, Minwoo Shin}
}

\title{CAFE+FNO: Fourier Kernel Generation via Multiplicative Feature Composition}

\author[1]{Hyungjoon Juen}
\author[1]{Minwoo Shin\thanks{Corresponding author}}

\affil[1]{Department of Software, Yonsei University (Mirae Campus), Wonju 26493, Republic of Korea}

\date{}

\begin{document}
\maketitle

\begin{abstract}
The Fourier Neural Operator (FNO) learns solution operators of partial differential equations (PDEs) through Fourier-space kernel parameterization, but frequency truncation can limit the learning of high-frequency variations. AM-FNO and SirenFNO generate kernels for all grid modes from spectral
coordinates using shared networks, making coordinate encoding and generator design important. Recent work on implicit neural representations (INRs) has proposed constructing frequency interactions through explicit feature composition rather than relying on subsequent MLPs to form them implicitly. Building on this approach, we propose CAFE+FNO, which incorporates Content-Aware Frequency Encoding+ (CAFE+) into Fourier kernel generation. CAFE+ combines Fourier--Chebyshev features through parallel affine branches and a Hadamard product, forming interactions within and across the two feature families. A kernel MLP maps the resulting representation of each normalized spectral coordinate to a complex channel-mixing matrix. Each layer shares its generator across all stored modes, making the number of trainable parameters independent of the number of modes for a fixed architecture. We compare CAFE+FNO with existing FNO variants on five PDE benchmarks and conduct ablation studies on basis configuration, multiplicative composition, and bandwidth learnability. Code and experimental configurations are available at
\url{https://github.com/fabsk101/CAFEPlusFNO.git}.
\end{abstract}

\section{Introduction}
\label{sec:introduction}
Partial differential equations (PDEs) describe the spatiotemporal behavior of physical systems, including fluid flow, heat transfer, and elastic deformation. Traditional numerical methods, such as the finite difference method (FDM) and the finite element method (FEM), provide reliable solutions but incur substantial costs in repeated analyses since the PDE must be solved again for each new initial or boundary condition. Neural operators approximate the mapping from an input function to a solution function, enabling much faster prediction than conventional numerical methods without being restricted to a particular data discretization. In particular, the Fourier Neural Operator (FNO) efficiently models global interactions by computing kernel integrals in the Fourier domain \citep{li2021fno}.

FNO assigns an independent channel-mixing matrix to each Fourier mode. Parameterizing all modes increases kernel parameter counts and memory costs with the number of modes and the problem dimension. Standard implementations therefore retain a prescribed number of low-frequency modes and discard the others \citep{george2024}.

Amortized Fourier Neural Operator (AM-FNO) and SirenFNO address this frequency truncation by generating full-frequency Fourier kernels with neural networks that take spectral coordinates as input, instead of storing each kernel coefficient independently \citep{xiao2024amfno,sirenfno2026}. In this amortized parameterization, each layer shares its kernel generator across all modes, keeping the number of learnable parameters independent of grid resolution. Consequently, kernel coefficients of each mode are outputs of a shared function $R(\boldsymbol{\xi}_k)$ rather than independent parameters.

This shifts the focus from storing mode-specific kernel coefficients to approximating a frequency-to-kernel function shared across all modes. Full-frequency kernel learning thus requires considering both mode coverage and the representation used to generate mode-dependent complex kernels from spectral coordinates.

To improve this representation, we propose CAFE+FNO, which incorporates the multiplicative feature composition of Content-Aware Frequency Encoding+ (CAFE+) \citep{ke2026cafe} into the amortized Fourier kernel generation of AM-FNO. CAFE+FNO constructs learnable interactions among spectral-coordinate features before kernel prediction and uses them to generate a complex channel-mixing kernel for each Fourier mode. It retains amortized parameterization over all stored modes without additional frequency truncation while introducing a learnable feature-composition stage before kernel prediction.

CAFE+FNO achieves the lowest mean error on four of five PDE benchmarks, reducing error relative to SirenFNO by 18.4--60.2\% on these tasks; SirenFNO performs better on ReacDiff.

\begin{enumerate}
  \item We propose CAFE+FNO, which introduces the Fourier--Chebyshev features and multiplicative composition of CAFE+ into spectral-coordinate-to-kernel mapping.
  \item We formulate CAFE+ based kernel generation using both dense parameterization and functional tensor parameterizations based on CP, TT, and Tucker decompositions.
  \item We compare against existing FNO variants under the same dataset-specific training and evaluation conditions and analyze the empirical effects of the architecture through ablations on the basis configuration, composition block, and bandwidth learning.
\end{enumerate}

\section{Related Work}
\label{sec:related-work}
\paragraph{Neural operators.}
Neural operators learn mappings between function spaces \citep{bhattacharya2021,kovachki2024}. DeepONet \citep{lu2021deeponet} combines branch and trunk networks, while graph-based neural operators learn kernels between irregularly sampled points \citep{li2020graph,li2020multipole}. Attention-based methods, including Galerkin Transformer \citep{cao2021}, OFormer \citep{li2023oformer},
GNOT \citep{hao2023gnot}, and Transolver \citep{wu2024transolver}, model interactions among discretized points or latent representations. FNO \citep{li2021fno} implements global convolution through mode-wise Fourier transformations, retaining a prescribed set of low-frequency modes. Subsequent work improves multiscale representations \citep{rahman2023}; U-FNO \citep{wen2022ufno} adds a U-Net path to complement high-frequency representations, while F-FNO \citep{tran2023ffno} and tensorized FNOs \citep{kossaifi2024} reduce costs by factorizing spectral operations and kernel tensors, respectively. AM-FNO \citep{xiao2024amfno} and SirenFNO \citep{sirenfno2026} generate kernels for all stored Fourier modes using shared networks conditioned on spectral coordinates. For fixed architectures, their parameter counts are independent of the number of modes. AM-FNO employs a KAN \citep{liu2025kan} or a Chebyshev-basis MLP, whereas SirenFNO uses a SIREN with learnable random Fourier features (RFFs). Our work introduces CAFE+ feature composition into this shared kernel parameterization.

\paragraph{Coordinate encodings for implicit neural representations.} 
Coordinate-based implicit neural representations (INRs) map coordinates to continuous signals, but conventional MLPs tend to learn low-frequency components first \citep{rahaman2019,xu2024}. Multiplicative Filter Networks (MFNs) multiply coordinate-dependent sinusoidal or Gabor features with linearly transformed hidden features at each layer \citep{fathony2021}. BACON constrains output bandwidths \citep{lindell2022}, while Residual MFNs combine skip connections with frequency initialization for multiscale learning \citep{shekarforoush2022}. SIREN uses sinusoidal activations \citep{sitzmann2020}, whereas Fourier feature mapping provides sinusoidal input embeddings \citep{tancik2020}. Fourier encodings can be sensitive to frequency selection and bandwidth \citep{hertz2021,benbarka2022,ma2025}. Nonlinear layers can generate harmonics and combinations of the encoded frequencies \citep{yuce2022}. CAFE introduces multiplicative feature composition, and CAFE+ adds Chebyshev features for more stable low-frequency representations \citep{ke2026cafe}. We use this encoding to generate Fourier kernels rather than directly reconstruct signals.

\section{Preliminaries}
\label{sec:preliminaries}
\subsection{Fourier Integral Operator}
\label{sec:fourier-integral-operator}

We consider the problem of approximating a solution operator between function spaces from a finite collection of input--output function pairs \citep{nelsen2021}. Let $\Omega \subset \mathbb{R}^{d_s}$ be a bounded open computational domain, let $\mathcal{A}$ denote the admissible set of input functions $a\colon\Omega\rightarrow\mathbb{R}^{d_a}$, and let $\mathcal{U}$ be a Banach space containing the solution functions $u\colon\Omega\rightarrow\mathbb{R}^{d_u}$. Here, $\mathcal{A}$ is a subset of a suitable Banach function space. Assuming that each input $a\in\mathcal{A}$ uniquely determines a solution $u\in\mathcal{U}$, we define the corresponding solution operator as
\begin{equation}
    \mathcal{G}^{\star}\colon\mathcal{A}\rightarrow\mathcal{U},
    \qquad u=\mathcal{G}^{\star}(a).
    \label{eq:prelim-solution-operator}
\end{equation}

Suppose that $N$ input--output function pairs $(a_i,u_i)$, $i=1,\ldots,N$, are available, where $u_i$ denotes observations of the solution corresponding to input $a_i$. Our goal is to use these observations to construct an operator $\mathcal{G}\colon\mathcal{A}\rightarrow\mathcal{U}$ that approximates $\mathcal{G}^{\star}$. To this end, we parameterize $\mathcal{G}$ using a neural network with finitely many trainable parameters and learn these parameters by reducing the discrepancy between the predictions $\mathcal{G}(a_i)$ and observations $u_i$.

To construct this approximation, FNO first maps the input function to a hidden function and then repeatedly applies layers consisting of an integral operator and a pointwise transformation.

Let $\mathcal{V}=L^2(\Omega;\mathbb{R}^{d_v})$ be the hidden function space, and denote the input to layer $\ell$ by $v^{(\ell)}\in\mathcal{V}$. Here, $d_v$ is the dimension of the hidden representation, which remains the same across all layers.

The Fourier representation below describes convolution on a periodic extension of a rectangular computational domain. This computational periodic representation is distinct from the physical boundary conditions of the underlying PDE. For an integrable matrix-valued kernel $\kappa_\ell$, the Fourier integral operator is given by
\begin{equation}
    \begin{aligned}
    (K_\ell v^{(\ell)})(x)
        &=\int_\Omega \kappa_\ell(x-y)v^{(\ell)}(y)\,\mathrm{d}y\\
        &=\mathcal{F}^{-1}\!\left(R_\ell\cdot\mathcal{F}[v^{(\ell)}]\right)(x),
    \end{aligned}
    \label{eq:prelim-fourier-integral}
\end{equation}
where $\mathcal{F}$ and $\mathcal{F}^{-1}$ denote the Fourier transform and inverse Fourier transform, respectively, and $R_\ell(k)=\mathcal{F}[\kappa_\ell](k)\in\mathbb{C}^{d_v\times d_v}$ is the complex-valued kernel matrix associated with Fourier mode $k$.

In the discrete implementation considered hereafter, we denote the finite index set of Fourier modes represented by the computational grid by $E\subset\mathbb{Z}^{d_s}$.

\subsection{CAFE+ Encoding}
\label{sec:cafe-encoding}

CAFE introduces learnable composition of Fourier features; CAFE+ extends this encoding with Chebyshev features. The concatenated features pass through parallel affine branches, whose outputs are multiplied elementwise before entering a backbone MLP. This constructs interactions within and across the two feature families. Originally designed for coordinate-based signal reconstruction, CAFE+ is used here to encode spectral coordinates for Fourier kernel generation. Section 4 defines the corresponding feature maps and composition. For an input coordinate $\mathbf{x}\in\mathbb{R}^{D}$ and $M$ Fourier frequency vectors $\{\boldsymbol{\omega}_i\}_{i=1}^{M}$, where $\boldsymbol{\omega}_i\in\mathbb{R}^{D}$, the Fourier features are defined as
\begin{equation}
    \Phi_{\mathrm{FF}}(\mathbf{x})
    =\left[\sin(2\pi\boldsymbol{\omega}_i^{\top}\mathbf{x}),\,
    \cos(2\pi\boldsymbol{\omega}_i^{\top}\mathbf{x})\right]_{i=1}^{M}.
    \label{eq:prelim-fourier-features}
\end{equation}
Here, $M$ is the number of Fourier frequency vectors, and $\boldsymbol{\omega}_i$ is a Fourier frequency vector.

Alongside the Fourier features, CAFE+ uses $J$ Chebyshev polynomials of the first kind $T_0,\ldots,T_{J-1}$ along each coordinate axis to construct the Chebyshev features:
\begin{equation}
    \Phi_{\mathrm{CF}}(\mathbf{x})
    =\left[T_j(x_d)\right]_{d=1,\ldots,D;\ j=0,\ldots,J-1}.
    \label{eq:prelim-chebyshev-features}
\end{equation}

CAFE+ concatenates the two sets of features, feeds them into $N_b$ linear branches, and combines the branch outputs using a Hadamard product:
\begin{equation}
    \Psi(\mathbf{x})
    =\bigodot_{i=1}^{N_b}
    \left\{\mathbf{W}_i\left[\Phi_{\mathrm{FF}}(\mathbf{x}),\,
    \Phi_{\mathrm{CF}}(\mathbf{x})\right]+b_i\right\},
    \label{eq:prelim-multiplicative-composition}
\end{equation}
where $\bigodot$ denotes Hadamard product.

This multiplicative composition establishes interactions between Fourier and Chebyshev bases at the encoding stage, and the resulting $\Psi(\mathbf{x})$ is subsequently used as the input to the backbone MLP.

\section{Method}
\label{sec:method}

We propose CAFE+FNO, which uses the multiplicative feature composition of CAFE+ as the input representation for a Fourier kernel generator. Whereas the original CAFE+ constructs a representation for reconstructing signal values from coordinates, our approach generates the kernel of each Fourier mode from normalized spectral coordinates. We first describe the dense kernel generator; extensions based on CP, TT, and Tucker factorizations are presented in Appendix~\ref{app:factorization}.

\begin{figure}[t]
    \centering
    \includegraphics[width=\linewidth]{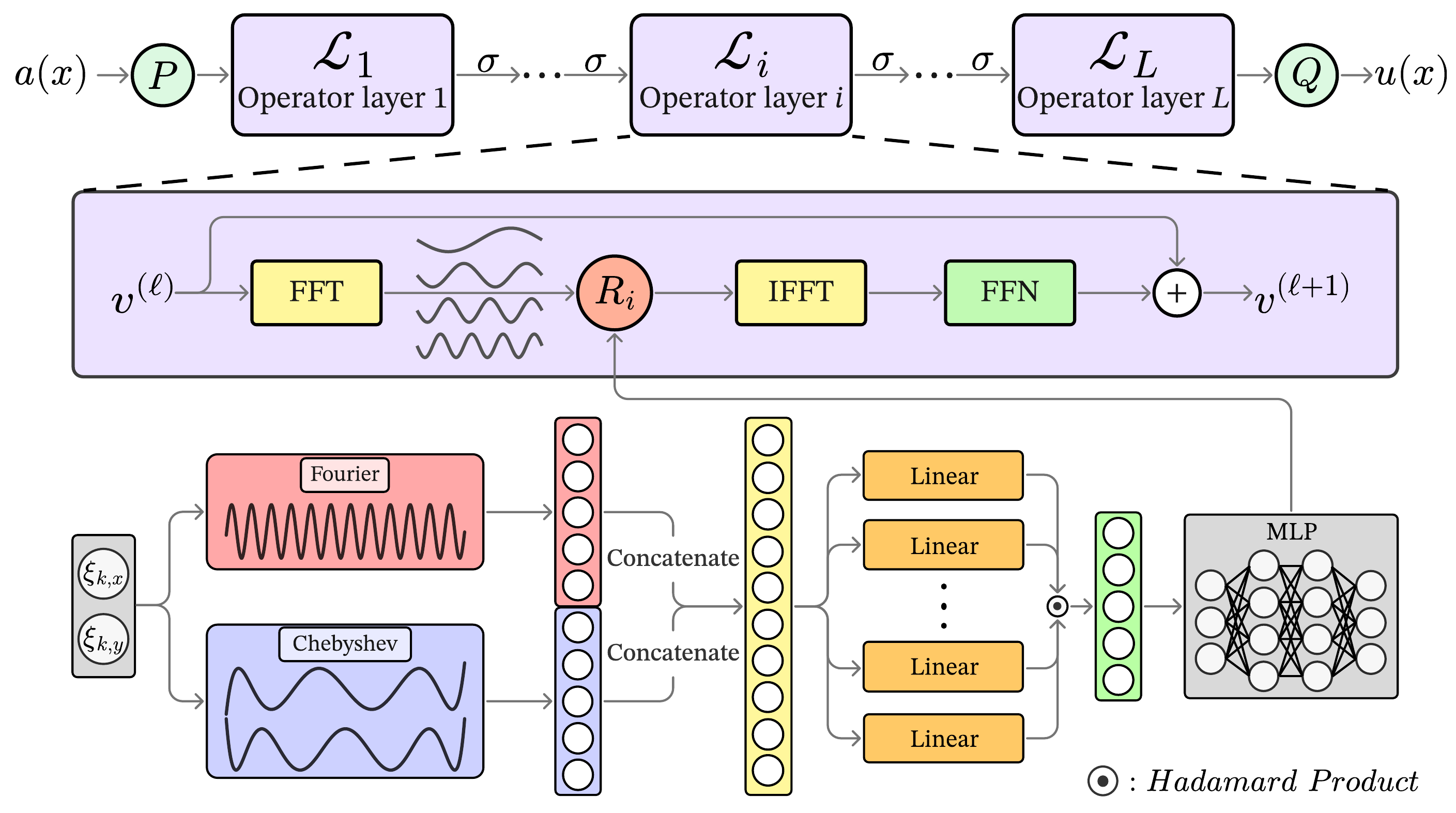}
    \caption{
    Architecture of CAFE+FNO in two dimensions.
    The input is lifted by $P$, processed by $L$ operator layers,
    and projected by $Q$ to the predicted solution.
    Within each layer, random Fourier features and Chebyshev features
    of the normalized spectral coordinate $\boldsymbol{\xi}_k$
    are concatenated and passed through parallel linear branches.
    Their outputs are combined via a Hadamard product,
    and a kernel MLP generates the mode-specific complex
    channel-mixing matrix used in the spectral convolution.
    }
    \label{fig:architecture}
\end{figure}

\subsection{Fourier--Chebyshev Feature Encoding}
\label{sec:feature_encoding}

Let $\boldsymbol{\xi}_k\in[-1,1]^{d_s}$ denote the normalized spectral coordinate corresponding to each Fourier mode $k\in E$. The coordinate $\boldsymbol{\xi}_k$ is the input to the kernel generator; the coordinate normalization is specified in Appendix~\ref{app:coordinates}. In each layer $\ell$, we independently sample the entries of a Gaussian matrix $G_\ell$ once from the standard normal distribution to map this coordinate to $m$ pairs of Fourier features:
\begin{equation}
G_\ell\in\mathbb{R}^{d_s\times m},
\qquad (G_\ell)_{ij}\sim\mathcal{N}(0,1).
\label{eq:gaussian_projection}
\end{equation}
Here, $d_s\in\{1,2\}$ is the spatial dimension. The sampled $G_\ell$ remains fixed during training, and a Fourier projection matrix $B_\ell=\rho_\ell G_\ell$ is constructed using $\rho_\ell>0$. The matrix $B_\ell\in\mathbb{R}^{d_s\times m}$ maps the normalized spectral coordinate to RFF phases in the $\ell$th operator layer.

Our implementation defines the RFF phase as $\pi B_\ell^\top\boldsymbol{\xi}_k$, with the frequency scale incorporated into $B_\ell$.

The RFF encoding of the spectral coordinate in layer $\ell$ is defined as
\begin{equation}
\Phi_{\mathrm{FF}}^{(\ell)}(\boldsymbol{\xi}_k;\rho_\ell)
=\begin{bmatrix}
\cos(\pi B_\ell^\top\boldsymbol{\xi}_k)\\
\sin(\pi B_\ell^\top\boldsymbol{\xi}_k)
\end{bmatrix}
\in\mathbb{R}^{2m}.
\label{eq:method_rff}
\end{equation}
Sine and cosine are applied elementwise, and their outputs are concatenated into a $2m$-dimensional feature vector. The parameter $\rho_\ell$ determines the frequency scale of the spectral-coordinate encoding by controlling the magnitude of the Gaussian projection.

Since CAFE+ uses Chebyshev features alongside Fourier features, we also compute $n$ first-kind Chebyshev basis functions for each coordinate axis. Writing $t_n(s)=[T_0(s),\ldots,T_{n-1}(s)]^\top\in\mathbb{R}^{n}$, we obtain
\begin{equation}
\Phi_{\mathrm{CF}}(\boldsymbol{\xi}_k)
=
\left[
t_n(\xi_{k,1})^\top,\,
\ldots,\,
t_n(\xi_{k,d_s})^\top
\right]^\top
\in\mathbb{R}^{d_s n}.
\label{eq:method_chebyshev}
\end{equation}
The Fourier and Chebyshev features are then concatenated:
\begin{equation}
\boldsymbol{\gamma}_\ell(\boldsymbol{\xi}_k)
=
\left[
\Phi_{\mathrm{CF}}(\boldsymbol{\xi}_k)^\top,\,
\Phi_{\mathrm{FF}}^{(\ell)}
(\boldsymbol{\xi}_k;\rho_\ell)^\top
\right]^\top.
\label{eq:combined_encoding}
\end{equation}
The resulting feature is $\boldsymbol{\gamma}_\ell(\boldsymbol{\xi}_k)\in\mathbb{R}^{2m+d_s n}$, with embedding dimension $2m+d_s n$.

\subsection{CAFE+ Multiplicative Frequency Composition}
\label{sec:frequency_composition}

CAFE+FNO feeds $\boldsymbol{\gamma}_\ell(\boldsymbol{\xi}_k)$ into $N_b$ parallel branches before passing the resulting representation to the kernel MLP. Each branch applies an affine transformation with a bias, referred to hereafter as a linear branch:
\begin{equation}
h_{\ell,r}(\boldsymbol{\xi}_k)
=\widetilde{W}_{\ell,r}\boldsymbol{\gamma}_\ell(\boldsymbol{\xi}_k)
+\widetilde{b}_{\ell,r},
\qquad r=1,\ldots,N_b,
\label{eq:linear_branch}
\end{equation}
where $\widetilde{W}_{\ell,r}\in\mathbb{R}^{d_z\times(2m+d_s n)}$ and $\widetilde{b}_{\ell,r}\in\mathbb{R}^{d_z}$. The branch outputs are combined by a Hadamard product:
\begin{equation}
\mathbf{z}_\ell(\boldsymbol{\xi}_k)
=\bigodot_{r=1}^{N_b}h_{\ell,r}(\boldsymbol{\xi}_k).
\label{eq:multiplicative_composition}
\end{equation}
Since each branch is an affine transformation of $\boldsymbol{\gamma}_\ell(\boldsymbol{\xi}_k)$, each component of $\mathbf{z}_\ell(\boldsymbol{\xi}_k)$ can be expressed as a polynomial of degree at most $N_b$ in the input feature components. In particular, two branches yield quadratic interactions together with linear and constant terms, whose coefficients are parameterized by products of branch weights and biases.

To examine the frequency structure of these interactions, we fix layer $\ell$ and write the Fourier projection matrix as $B_\ell=[\boldsymbol{\omega}_{\ell,1},\ldots,\boldsymbol{\omega}_{\ell,m}]$. Here, $\boldsymbol{\omega}_{\ell,i}$ is an encoding frequency vector for the spectral coordinate $\boldsymbol{\xi}_k$, distinct from the Fourier mode $k$ of the hidden function. We define the phase associated with each column vector as $\theta_{\ell,i}=\pi\boldsymbol{\omega}_{\ell,i}^\top\boldsymbol{\xi}_k$ for $i=1,\ldots,m$.
For this example, we abbreviate $\theta_i:=\theta_{\ell,i}$ and $\theta_j:=\theta_{\ell,j}$. We simplify the same component of the two branch output vectors as
\begin{equation}
h_1=\alpha_1\cos\theta_i+\beta_1\sin\theta_i,
\qquad h_2=\alpha_2\cos\theta_j+\beta_2\sin\theta_j.
\label{eq:two_branch_example}
\end{equation}
Applying the product-to-sum identities gives
\begin{equation}
\begin{aligned}
h_1h_2=\tfrac{1}{2}\big[{}
 &(\alpha_1\alpha_2+\beta_1\beta_2)\cos(\theta_i-\theta_j)
 +(\alpha_1\alpha_2-\beta_1\beta_2)\cos(\theta_i+\theta_j)\\
 &+(\beta_1\alpha_2-\alpha_1\beta_2)\sin(\theta_i-\theta_j)
 +(\beta_1\alpha_2+\alpha_1\beta_2)\sin(\theta_i+\theta_j)\big].
\end{aligned}
\label{eq:product_to_sum}
\end{equation}
Since
\begin{equation}
\theta_i\pm\theta_j
=\pi(\boldsymbol{\omega}_{\ell,i}\pm\boldsymbol{\omega}_{\ell,j})^\top
\boldsymbol{\xi}_k,
\label{eq:sum_difference_phase}
\end{equation}
the representation after the Hadamard product can include, in addition to the initial RFF basis, basis frequencies corresponding to the sums $\boldsymbol{\omega}_{\ell,i}+\boldsymbol{\omega}_{\ell,j}$ and differences $\boldsymbol{\omega}_{\ell,i}-\boldsymbol{\omega}_{\ell,j}$. The coefficient of each component is determined by the learnable linear-branch weights $\alpha_1,\beta_1,\alpha_2,\beta_2$, allowing the network to selectively adjust the frequency interactions needed for kernel generation.

Products of Chebyshev components similarly satisfy
\begin{equation}
T_p(x)T_q(x)=\tfrac{1}{2}\big[T_{p+q}(x)+T_{|p-q|}(x)\big],
\qquad 0\leq p,q\leq n-1.
\label{eq:chebyshev_product}
\end{equation}
Thus, multiplying Chebyshev features along the same coordinate axis can construct higher-order polynomial components absent from the initial encoding, while also recombining lower-order components.

Furthermore, since both feature types enter the same branches, products of branch outputs can contain Fourier--Chebyshev cross terms such as $T_p(\xi_{k,a})\cos\theta_{\ell,i}$ and $T_p(\xi_{k,a})\sin\theta_{\ell,i}$. These terms form sinusoidal features whose amplitudes are modulated by polynomials. Consequently, $\mathbf{z}_\ell(\boldsymbol{\xi}_k)$ represents both interactions within each basis and cross interactions between the two bases. The kernel MLP in the next subsection uses this representation to generate a complex-valued kernel matrix for each mode.

The frequencies composed here are encoding frequencies with respect to the spectral coordinate $\boldsymbol{\xi}_k$. This operation neither directly multiplies different Fourier coefficients of the PDE solution nor adds new Fourier modes to the computational grid. The purpose of the operation is to provide basis interactions for representing the mapping from $\boldsymbol{\xi}_k$ to the kernel matrix. Its effect on solution prediction is evaluated in the experiments in Section~\ref{sec:experiments}.

\subsection{Complex Fourier Kernel Generation}
\label{sec:complex_kernel}

Let $v^{(\ell)}:\Omega\rightarrow\mathbb{R}^{d_v}$ be the hidden representation at layer $\ell$. The CAFE+ mode representation $\mathbf{z}_\ell(\boldsymbol{\xi}_k)\in\mathbb{R}^{d_z}$ constructed above is passed to a complex matrix-valued kernel generator $\mathcal{M}_\ell:\mathbb{R}^{d_z}\rightarrow\mathbb{C}^{d_v\times d_v}$. In practice, we use a single real-valued MLP $f_\ell:\mathbb{R}^{d_z}\rightarrow\mathbb{R}^{2d_v^2}$. Its first and last $d_v^2$ output components are split into real and imaginary parts, respectively, and reshaped into $d_v\times d_v$ matrices:
\begin{equation}
\begin{bmatrix}
r_\ell^{\mathrm{Re}}(k)\\
r_\ell^{\mathrm{Im}}(k)
\end{bmatrix}
=f_\ell\big(\mathbf{z}_\ell(\boldsymbol{\xi}_k)\big),
\qquad r_\ell^{\mathrm{Re}}(k),r_\ell^{\mathrm{Im}}(k)\in\mathbb{R}^{d_v^2}.
\label{eq:real_imaginary_outputs}
\end{equation}
The kernel corresponding to Fourier mode $k$ is therefore defined as
\begin{equation}
R_\ell(k)=\mathcal{M}_\ell\big(\mathbf{z}_\ell(\boldsymbol{\xi}_k)\big)
=R_\ell^{\mathrm{Re}}(k)+\mathrm{i}R_\ell^{\mathrm{Im}}(k).
\label{eq:generated_kernel}
\end{equation}
The same $\mathcal{M}_\ell$ and encoding parameters are shared across all Fourier modes within layer $\ell$, whereas different layers use independent parameters.

The kernel generator takes only spectral coordinates as input and is not conditioned on an individual input function $a$. Thus, after training, the kernel generated for a given mode on the same grid and in the same layer is identical across input samples. Here, content-aware composition means that end-to-end training through the PDE solution prediction error forms feature combinations suited to the data. It does not imply supervision with separate ground-truth kernels or the generation of a different kernel for each input sample.

Consequently, when the generator architecture and feature dimensions are fixed, the number of trainable parameters is independent of the size of the mode set $E$. However, the cost of evaluating and storing kernels over all modes increases with the grid size. We apply a Hermitian projection to the generated complex kernels to maintain consistency with the inverse Fourier transform of a real-valued hidden function:
\begin{equation}
\widetilde{R}_\ell=\Pi_{\mathcal{H}}(R_\ell),
\qquad \widetilde{R}_\ell(-k)=\overline{\widetilde{R}_\ell(k)}.
\label{eq:hermitian_projection}
\end{equation}
This symmetry condition applies to the completed spectrum. The specific projection used for the rFFT storage array is described in Appendix~\ref{app:coordinates}.

Finally, $\widetilde{R}_\ell(k)$ acts linearly on the Fourier coefficients of the hidden representation at each mode:
\begin{equation}
\mathcal{F}[K_\ell v^{(\ell)}](k)
=(\widetilde{R}_\ell\cdot\widehat{v}^{(\ell)})(k),
\qquad \widehat{v}^{(\ell)}=\mathcal{F}[v^{(\ell)}].
\label{eq:kernel_action}
\end{equation}

\subsection{CAFE+FNO Architecture}
\label{sec:architecture}

CAFE+FNO follows the basic operator architecture of AM-FNO, replacing only the kernel generator in each Fourier layer with the CAFE+ based parameterization defined above. The complete model is written as
\begin{equation}
\mathcal{G}:=Q\circ\mathcal{L}_L\circ\mathcal{L}_{L-1}\circ\cdots\circ\mathcal{L}_1\circ P,
\label{eq:operator_architecture}
\end{equation}
with $v^{(1)}=P(a)$ and $\widehat{u}=Q\big(v^{(L+1)}\big)$. The lifting operator $P:\mathcal{A}\rightarrow\mathcal{V}$ constructs a $d_v$-dimensional hidden function from the values of the input function and the computational coordinates. Each operator layer $\mathcal{L}_\ell$ updates the hidden function, and the projection operator $Q:\mathcal{V}\rightarrow\mathcal{U}$ maps the final hidden function to the predicted solution function. 

Let $\widetilde{R}_\ell(k)$ denote the generated complex kernel after Hermitian projection. The $\ell$th layer is computed as
\begin{equation}
\begin{aligned}
\widetilde{v}^{(\ell)}
&=\mathcal{F}^{-1}\!\left(\widetilde{R}_\ell\cdot\mathcal{F}[v^{(\ell)}]\right),\\
v^{(\ell+1)}(x)
&=\sigma\!\left(v^{(\ell)}(x)+\mathrm{FFN}_\ell\big(\widetilde{v}^{(\ell)}(x)\big)\right).
\end{aligned}
\label{eq:operator_layer}
\end{equation}
Here, $\mathrm{FFN}_\ell(y)=\widehat{W}_\ell^{(2)}\sigma(\widehat{W}_\ell^{(1)}y+b_\ell^{(1)})+b_\ell^{(2)}$, and $\sigma$ is the activation function. In the final operator layer, only the activation after the residual addition is omitted; the activation within the FFN is retained. The final hidden representation is mapped to the output through the same projection.

\section{Experiments}
\label{sec:experiments}
This section presents the experimental results. All experiments were conducted on a desktop computer equipped with an AMD Ryzen 9800XD 8-Core CPU, an NVIDIA GeForce RTX 5080 GPU, and 64GB of Samsung DDR5-5600 RAM (16GB x 4).

\subsection{Benchmarks and baselines}
\label{sec:benchmarks}
We compare CAFE+FNO with FNO and several established variants: U-FNO, TFNO-CP with CP decomposition, AM-FNO with an MLP, and SirenFNO with a SIREN hypernetwork. We also compare CP-, TT-, and Tucker-based variants of both SirenFNO and CAFE+FNO.

Table~\ref{tab:datasets} lists the benchmark datasets. Darcy flow \citep{neuraloperator_darcy_data} and Navier--Stokes \citep{neuraloperator_ns_data} are obtained from the NeuralOperator library; Burgers and reaction--diffusion are from PDEBench \citep{takamoto2024pdebench,pdebench_data}; and Airfoil is from the Geo-FNO benchmark \citep{li2024geofno}.

For Airfoil, the publicly available materials were insufficient to reconstruct the corresponding U-FNO implementation consistently, and the fixed baseline implementation required additional compatibility modifications. We therefore exclude U-FNO from this benchmark and indicate the corresponding entry with a dash in Table~\ref{tab:results}.

\begin{table}[t]
\caption{Description of the benchmark datasets.}
\label{tab:datasets}
\begin{center}
\begin{tabularx}{\textwidth}{@{}lXXcll@{}}
\toprule
Dataset & Input & Target & Resolution & Train/test & Source \\
\midrule
Darcy & $a(x)$ & $u(x)$ & $128\!\times\!128$ & 1000/200 & NeuralOperator \\
NS & $u(x,t_{\mathrm{in}})$ & $u(x,t_{\mathrm{out}})$ & $128\!\times\!128$ & 1000/200 & NeuralOperator \\
Burgers & $[u(x,t_j)]_{j=0}^{9}$ & $[u(x,t_j)]_{j=10}^{19}$ & $1024$ & 1000/200 & PDEBench \\
Airfoil & $(X,Y)$ & $u(x)$ & $221\!\times\!51$ & 1000/200 & Geo-FNO \\
ReacDiff & $[u(x,t_j)]_{j=0}^{9}$ & $[u(x,t_j)]_{j=10}^{19}$ & $1024$ & 1000/200 & PDEBench \\
\bottomrule
\end{tabularx}
\end{center}
\end{table}

Appendix~\ref{app:data} provides dataset-specific preprocessing,
prediction protocols, training losses, and evaluation settings.

\subsection{Experimental setup}
\label{sec:setup}
For a fair comparison, we retrained FNO, U-FNO, TFNO-CP, AM-FNO, SirenFNO, and CAFE+FNO using the same data splits, preprocessing, and training pipeline. Each baseline architecture follows the dataset-specific settings of its public implementation. For Airfoil, which is not included in the original SirenFNO benchmarks, we used a separately specified architecture configuration. 

All models were trained for 500 epochs using AdamW, with an initial learning rate of $10^{-3}$ and weight decay of $10^{-4}$. The learning rate was reduced using cosine annealing. We used neither validation-based model selection nor early stopping; all results were evaluated using the final-epoch model. The main evaluation metric is the mean sample-wise relative $L_2$ error,
\begin{equation}
\mathcal{L}_{\mathrm{rel}}=
\frac{1}{N}\sum_{i=1}^{N}
\frac{\lVert\mathcal{G}(a_i)-u_i\rVert_2}{\lVert u_i\rVert_2}.
\label{eq:relative-error}
\end{equation}
Here, $N$ is the number of evaluation samples, and $u_i$ and $\mathcal{G}(a_i)$ are the targets and predictions expressed in the dataset-specific evaluation space. For Burgers and ReacDiff, sample-wise relative $L_2$ errors are computed by concatenating the temporal and spatial values over the entire prediction rollout. The other benchmarks are evaluated on the predicted target fields. 

CAFE+FNO, SirenFNO, and AM-FNO generate Fourier kernels for all stored FFT modes, while FNO and TFNO-CP use the frequency truncation of their respective public configurations. Each CAFE+FNO layer uses an independent Gaussian matrix $G_\ell$. For Table~\ref{tab:results}, $\rho_\ell$ was fixed to one throughout training. The effect of learning the bandwidth is examined in Section~\ref{sec:ablations}, and the CP, TT, and Tucker results are reported in Section~\ref{sec:main-results}.

Each experiment was repeated with five random seeds, $\{0,42,73,108,202\}$. We report the mean $\pm$ standard deviation of the final test errors. Dataset-specific batch sizes, preprocessing, training losses, and evaluation protocols are provided in Appendix~\ref{app:data}, CAFE+FNO architecture settings in Appendix~\ref{app:settings}, and complete configurations in the accompanying code.

\begin{table}[t]
\caption{Final test relative $L_2$ errors ($\times10^{-3}$) at epoch 500:
mean $\pm$ sample standard deviation over five seeds.
Bold marks the lowest mean per dataset.
AM-FNO uses the MLP variant; U-FNO was excluded a priori on Airfoil (---).
Evaluation details appear in Appendix~\ref{app:data}.}
\label{tab:results}

\begin{center}
\setlength{\tabcolsep}{4pt}

\begin{tabular*}{\textwidth}{
  @{\extracolsep{\fill}}l
  *{5}{
    r@{\extracolsep{0pt}\,{\color{black!65}$\pm$}\,}
    >{\color{black!65}}l@{\extracolsep{\fill}}
  }
  @{}
}
\toprule
Model
& \multicolumn{2}{c}{\textbf{Darcy}}
& \multicolumn{2}{c}{\textbf{NS}}
& \multicolumn{2}{c}{\textbf{Burgers}}
& \multicolumn{2}{c}{\textbf{Airfoil}}
& \multicolumn{2}{c}{\textbf{ReacDiff}} \\
&
\multicolumn{2}{c}{$128\times128$}
& \multicolumn{2}{c}{$128\times128$}
& \multicolumn{2}{c}{$1024$}
& \multicolumn{2}{c}{$221\times51$}
& \multicolumn{2}{c}{$1024$} \\
\midrule

FNO
& 65.40 & 4.18
& 56.29 & 0.93
& 11.30 & 0.08
& 6.17 & 0.17
& 4.18 & 0.15 \\

U-FNO
& 47.05 & 1.83
& 57.69 & 0.78
& 10.02 & 0.21
& \multicolumn{2}{c}{---}
& 9.13 & 5.77 \\

TFNO-CP
& 47.34 & 1.28
& 35.14 & 0.96
& 10.27 & 0.49
& 7.06 & 0.48
& 4.07 & 0.26 \\

AM-FNO(MLP)
& 62.02 & 12.90
& 29.94 & 1.83
& 18.73 & 3.20
& 6.29 & 0.68
& 1758.09 & 3843.63 \\

SirenFNO
& 24.34 & 1.24
& 43.24 & 2.02
& 5.86 & 0.41
& 5.76 & 0.24
& \textbf{2.36} & 0.78 \\

CAFE+FNO
& \textbf{13.70} & 0.13
& \textbf{17.20} & 0.32
& \textbf{4.13} & 0.37
& \textbf{4.70} & 0.12
& 3.80 & 0.27 \\

\midrule
\multicolumn{11}{@{}l}{\textit{Factorized SirenFNO}} \\

\quad CP
& 46.09 & 21.71
& 51.46 & 7.29
& 8.91 & 0.29
& 8.12 & 1.68
& 4.37 & 1.70 \\

\quad TT
& 178.78 & 257.76
& 36.04 & 5.53
& 7.72 & 0.21
& 6.02 & 0.26
& 2.63 & 0.23 \\

\quad Tucker
& 64.06 & 27.05
& 46.70 & 12.24
& 8.11 & 0.43
& 8.18 & 3.81
& 2.92 & 0.09 \\

\midrule
\multicolumn{11}{@{}l}{\textit{Factorized CAFE+FNO}} \\

\quad CP
& 30.05 & 2.57
& 68.91 & 26.94
& 8.73 & 0.84
& 8.68 & 2.52
& 4.15 & 0.48 \\

\quad TT
& 27.77 & 1.82
& 42.50 & 12.94
& 8.48 & 0.45
& 7.06 & 0.61
& 4.09 & 0.42 \\

\quad Tucker
& 27.96 & 2.22
& 39.58 & 5.92
& 8.97 & 0.47
& 8.13 & 1.22
& 3.72 & 0.54 \\

\bottomrule
\end{tabular*}
\end{center}
\end{table}

\subsection{Main results}
\label{sec:main-results}
Table~\ref{tab:results} shows that dense CAFE+FNO achieves the
lowest mean error on Darcy, NS, Burgers, and Airfoil.
Relative to SirenFNO, errors decrease by 43.7\%, 60.2\%, 29.5\%,
and 18.4\%, respectively. On ReacDiff, SirenFNO performs better
(2.36 versus 3.80, $\times10^{-3}$). CAFE+FNO also improves on
FNO and AM-FNO across all five datasets, although AM-FNO's
ReacDiff results show substantial variability.

Relative to dense CAFE+FNO, CP, TT, and Tucker variants reduce
parameter counts by 76.7--86.1\%, 68.7--74.7\%, and 0.5--78.0\%,
respectively (Table~\ref{tab:parameters}). Factorized CAFE+FNO variants have higher mean errors than the
dense model, except for Tucker on ReacDiff. All three outperform
their SirenFNO counterparts on Darcy; results elsewhere are mixed.
To target comparable parameter budgets, we use a 12-dimensional
input to each CAFE+ factor-generator MLP, compared with 16 or 32
in the corresponding SirenFNO models. Relative to dense CAFE+FNO,
both the initial basis counts and the MLP input dimension are
reduced. These changes may also affect accuracy; their effects
are not isolated from tensor factorization in this comparison.

\begin{figure}[htbp]
    \centering
    \includegraphics[width=\linewidth]{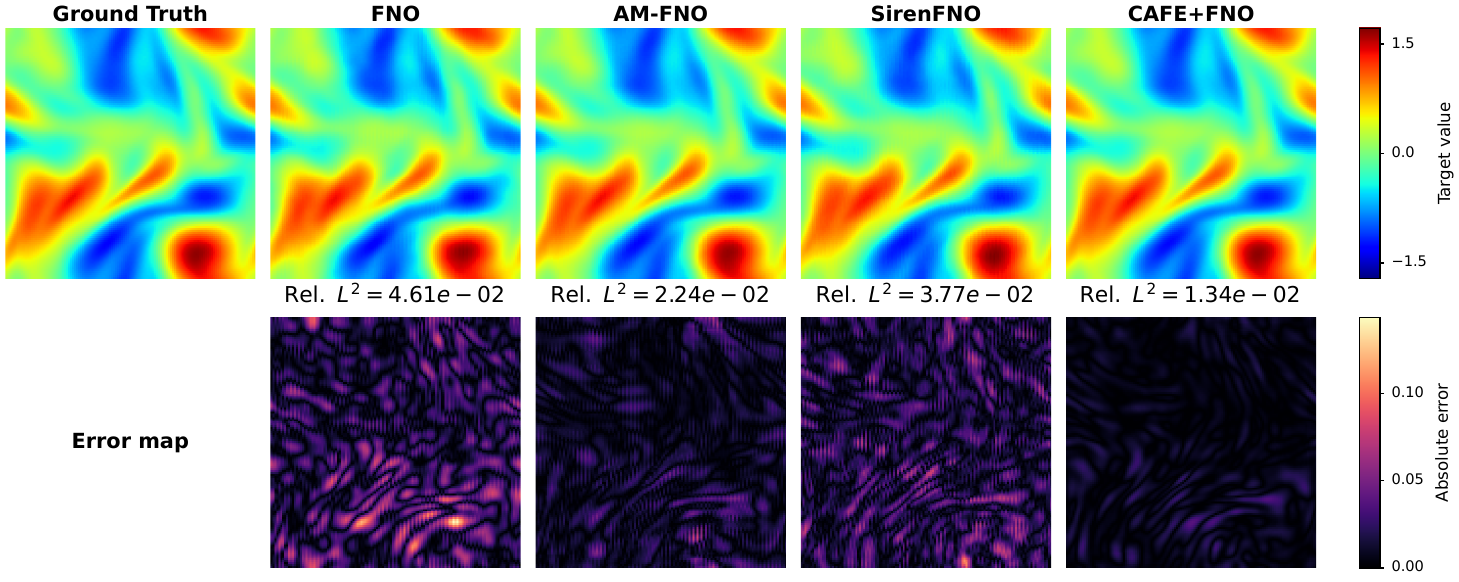}
    \caption{Comparison of FNO, AM-FNO, SirenFNO, and CAFE+FNO
    on a Navier--Stokes (NS) sample using trained checkpoints.
    The top row shows the ground-truth vorticity and model
    predictions; the bottom row shows pointwise absolute errors.
    Relative $L_2$ errors are reported below the predictions.}
    \label{fig:ns_prediction_comparison}
\end{figure}

\subsection{Ablation experiments}
\label{sec:ablations}
We analyze how the combination of Fourier and Chebyshev features, multiplicative feature composition through linear branches and a Hadamard product, and RFF bandwidth learnability affect prediction accuracy. Results are presented in Table~\ref{tab:ablation}. Since removing components may change feature dimensions and parameter counts, each comparison is interpreted as the empirical effect of the corresponding overall configuration.

\begin{table}[t]
\caption{CAFE+FNO ablations: relative $L_2$ errors ($\times10^{-3}$).
Evaluation and reporting conventions follow Table~\ref{tab:results}.
Bold marks the lowest unrounded mean per dataset.
$^{*}$: no composition block, with the kernel MLP input dimension
matched to the default model (16 for ReacDiff and 32 otherwise).
RFF/Cheb: single-basis variants retaining the composition block.
$^{\dagger}$: learnable bandwidth with a fixed Gaussian matrix.}
\label{tab:ablation}

\begin{center}
\setlength{\tabcolsep}{4pt}

\begin{tabular*}{\textwidth}{
  @{\extracolsep{\fill}}l
  *{5}{
    r@{\extracolsep{0pt}\,{\color{black!65}$\pm$}\,}
    >{\color{black!65}}l@{\extracolsep{\fill}}
  }
  @{}
}
\toprule
Model
& \multicolumn{2}{c}{\textbf{Darcy}}
& \multicolumn{2}{c}{\textbf{NS}}
& \multicolumn{2}{c}{\textbf{Burgers}}
& \multicolumn{2}{c}{\textbf{Airfoil}}
& \multicolumn{2}{c}{\textbf{ReacDiff}} \\
&
\multicolumn{2}{c}{$128\times128$}
& \multicolumn{2}{c}{$128\times128$}
& \multicolumn{2}{c}{$1024$}
& \multicolumn{2}{c}{$221\times51$}
& \multicolumn{2}{c}{$1024$} \\
\midrule

CAFE+FNO
& \textbf{13.63} & 0.12
& 17.25 & 0.29
& \textbf{4.11} & 0.37
& \textbf{4.69} & 0.10
& 3.80 & 0.27 \\

CAFE+FNO$^{*}$
& 27.28 & 2.70
& 38.53 & 1.46
& 5.14 & 0.34
& 17.58 & 11.38
& 3.87 & 0.28 \\

CAFE+FNO$_{\mathrm{RFF}}$
& 15.07 & 0.71
& 19.76 & 0.20
& 4.91 & 0.39
& 5.36 & 0.22
& 4.48 & 1.14 \\

CAFE+FNO$_{\mathrm{Cheb}}$
& 14.46 & 0.71
& 17.81 & 0.14
& 4.16 & 0.24
& 4.87 & 0.30
& 4.00 & 0.24 \\

CAFE+FNO$^{\dagger}$
& 13.69 & 0.14
& \textbf{17.22} & 0.26
& 4.13 & 0.37
& 4.69 & 0.12
& \textbf{3.65} & 0.29 \\

\bottomrule
\end{tabular*}
\end{center}
\end{table}

The combined Fourier--Chebyshev encoding yields lower mean errors
than either single-basis variant on all five datasets, supporting
its effectiveness for Fourier kernel generation.

With matched kernel MLP input dimensions (16 for ReacDiff and
32 otherwise), the default model achieves lower mean errors than
CAFE+FNO$^{*}$, which directly uses Fourier--Chebyshev features.
This supports the utility of CAFE+ encoding at the same MLP
input dimension.
Finally, CAFE+FNO$^{\dagger}$, with learnable RFF bandwidth
$\rho_\ell$, yields lower mean errors than the fixed-bandwidth
default on NS and ReacDiff, but higher errors on Darcy and Burgers.
Airfoil errors are equal at the reported precision.

\section{Conclusion}
\label{sec:conclusion}
We proposed CAFE+FNO, which introduces explicit, learnable feature composition into Fourier kernel generation. Its Fourier--Chebyshev encoding uses parallel affine branches and a Hadamard product to construct interactions within and across feature families before the kernel MLP. The dense model achieves the lowest mean error on four of five benchmarks, while SirenFNO performs better on ReacDiff. Ablations favor the combined basis over either basis alone and show lower mean errors with composition than with direct features at matched kernel MLP input dimensions. Together, these results support explicit spectral-coordinate feature composition as an effective design choice for Fourier kernel generation. Factorized variants use fewer parameters than dense CAFE+FNO, generally at some cost in accuracy. However, this comparison does not isolate tensor factorization from the accompanying reductions in basis counts and MLP input dimensions. Future work will focus on extending CAFE+FNO to a broader range of PDE problems and higher-dimensional spatial domains.

\label{lastmainpage}
\clearpage

\bibliography{references}
\bibliographystyle{unsrtnat}
\clearpage
\appendix

\section{Data Preprocessing, Training, and Evaluation}
\label{app:data}

\subsection{Data Splits and Preprocessing}

The data sources, splits, and input--output definitions follow Table~\ref{tab:datasets} in the main text. For Darcy, only the outputs are standardized using training-set statistics. For Navier--Stokes, the inputs and outputs are standardized separately using their respective training-set statistics. In both cases, the output normalization is inverted for evaluation. For Burgers, the global mean and population standard deviation are computed over all temporal and spatial values of 200 training trajectories selected separately for each seed. These statistics are used to normalize both inputs and outputs, and errors are computed in the normalized space.

\paragraph{Navier--Stokes.}
For Navier--Stokes, we use the input--output pairs in \texttt{nsforcing\_train\_128.pt} and \texttt{nsforcing\_test\_128.pt} provided by NeuralOperator. Each input and output is a single-channel vorticity field at a different time, and the corresponding output vorticity field is predicted directly from the input vorticity field. We do not use autoregressive rollout with multiple input snapshots in this experiment.

\paragraph{Airfoil.}
For Airfoil, the $X,Y$ coordinates of a $221\times51$ structured mesh are used as inputs to predict the Mach number at each grid point. The prediction target is the fifth component of the original flow-field array. No additional value normalization is applied to the input coordinates or output values.

\paragraph{Burgers and ReacDiff.}
For both benchmarks, we use the first 20 snapshots of each trajectory without temporal or spatial subsampling. Denoting the stored time points in the original data by $t_j$, the 10 snapshots at $t_0,\ldots,t_9$ are used as inputs, and the 10 snapshots at $t_{10},\ldots,t_{19}$ are predicted autoregressively. Thus, the input interval is $[t_0,t_9]$ and the prediction interval is $[t_{10},t_{19}]$, with the time step determined by the storage interval of the original data.

\subsection{Training and Evaluation Settings}
\begin{table}[!htbp]
\caption{Training and evaluation settings.}
\label{tab:training-evaluation}
\begin{center}
\begin{tabularx}{\textwidth}{@{}l c >{\raggedright\arraybackslash}X >{\raggedright\arraybackslash}p{0.20\textwidth}@{}}
\toprule
\textbf{Dataset}
& \shortstack{\textbf{Batch}\\\textbf{size}}
& \textbf{Prediction protocol}
& \textbf{Evaluation space} \\
\midrule
Darcy & 32
& Direct prediction of the provided input--output field pairs
& After output denormalization \\
Navier--Stokes & 32
& Direct prediction of the provided single input--output field pairs
& After output denormalization \\
Burgers & 32
& Autoregressive prediction of 10 future snapshots from 10 input snapshots
& Normalized space \\
Airfoil & 8
& Direct prediction of the target field from coordinates
& Original value space \\
ReacDiff & 32
& Autoregressive prediction of 10 future snapshots from 10 input snapshots
& Original value space \\
\bottomrule
\end{tabularx}
\end{center}
\end{table}

\paragraph{Training loss.}
For Darcy, Navier--Stokes, and Airfoil, training uses
sample-wise relative $L_2$ errors on the directly predicted
target fields. For Burgers and ReacDiff, the training loss is the sum of the relative $L_2$ errors over the 10 prediction steps. At each step, the per-sample relative errors are summed over the batch for Burgers and averaged over the batch for ReacDiff. For both benchmarks, previous predictions are incorporated into the input to the next step, and the computational graph is retained over the entire rollout for backpropagation during training. The Burgers loss is computed in the normalized space, whereas the ReacDiff loss is computed without value normalization.

\section{Frequency Coordinate Normalization}
\label{app:coordinates}

We use the Fourier mode $k$ and normalized spectral coordinate $\boldsymbol{\xi}_k$ defined in the main text. Let $L_a$ denote the length of the $a$th FFT axis after padding. The one-dimensional rFFT in the dense model stores nonnegative Fourier modes, with normalized coordinates given by
\begin{equation}
\boldsymbol{\xi}_k=\frac{2k}{L_1},\qquad
k=0,\ldots,\left\lfloor\frac{L_1}{2}\right\rfloor.
\end{equation}

Let $p=0,\ldots,L_1-1$ and $q=0,\ldots,\lfloor L_2/2\rfloor$ denote the integer storage indices of a two-dimensional rFFT. The components of the Fourier mode $k=(k_1,k_2)$ corresponding to storage position $(p,q)$ are defined as
\begin{equation}
k_1=
\begin{cases}
p, & 0\leq p\leq\left\lfloor(L_1-1)/2\right\rfloor,\\
p-L_1, & \left\lfloor(L_1-1)/2\right\rfloor<p<L_1,
\end{cases}
\qquad k_2=q.
\end{equation}

The first axis is therefore converted to signed frequency indices, while the last axis uses nonnegative frequency indices. The normalized coordinates are then given by
\begin{equation}
\boldsymbol{\xi}_k=
\left(\frac{2k_1}{L_1},\frac{2k_2}{L_2}\right)^{\top}.
\end{equation}

For the CP, TT, and Tucker models, midpoint coordinates are used for each input-component, output-component, and frequency-storage axis. For a factor axis of length $I$ with storage index $\nu$, the input coordinate to its factor generator is
\begin{equation}
s_I(\nu)=\frac{2(\nu+1/2)}{I}-1,\qquad
\nu=0,\ldots,I-1.
\end{equation}
Even on a frequency axis, $\nu$ follows the storage order and is therefore distinct from the signed frequency index used by the dense model. Hermitian projection is applied according to the rFFT storage convention. In one dimension, the kernels at the DC component and, for an even axis length, the Nyquist component are made real. In two dimensions, for the DC column of the last frequency axis and its Nyquist column when that axis has even length, each kernel along the first axis is averaged with the complex conjugate of the kernel at the opposite frequency. The remaining stored modes are unchanged, and the unstored modes are determined by conjugate symmetry. This operation is applied elementwise to each kernel matrix, without transposing the input- and output-component axes.

\section{CP, TT, and Tucker Tensor Decompositions}
\label{app:factorization}

Following the decomposition structures in Eqs.~(9)--(11) of Section~4.2 of SirenFNO~\citep{sirenfno2026}, we apply CAFE+ encoding to the factor generator for each axis. Let $d_s$ denote the spatial dimension, and arrange the $n_{\mathrm{ax}}=d_s+2$ axes in the order of the input components, the $d_s$ frequency axes, and the output components. The real and imaginary parts, indexed by $c\in\{\mathrm{R},\mathrm{I}\}$, are generated independently and combined as $U=U^{(\mathrm{R})}+iU^{(\mathrm{I})}$:
\begin{align}
\text{CP:}\quad
U^{(c)}
&=\sum_{\alpha=1}^{r}
\mathbf{f}_{1,\alpha}^{(c)}\otimes\cdots\otimes
\mathbf{f}_{n_{\mathrm{ax}},\alpha}^{(c)},\\
\text{TT:}\quad
U_{\nu_1,\ldots,\nu_{n_{\mathrm{ax}}}}^{(c)}
&=A_1^{(c)}(\nu_1)\cdots
A_{n_{\mathrm{ax}}}^{(c)}(\nu_{n_{\mathrm{ax}}}),\\
\text{Tucker:}\quad
U^{(c)}
&=\mathcal{C}^{(c)}\times_1F_1^{(c)}\cdots
\times_{n_{\mathrm{ax}}}F_{n_{\mathrm{ax}}}^{(c)}.
\end{align}

For tensor-axis lengths $I_a$, we have
$U^{(c)}\in\mathbb{R}^{I_1\times\cdots\times I_{n_{\mathrm{ax}}}}$.
The CP factors satisfy $\mathbf{f}_{a,\alpha}^{(c)}\in\mathbb{R}^{I_a}$,
the Tucker factor matrices satisfy $F_a^{(c)}\in\mathbb{R}^{I_a\times r}$,
and the trainable Tucker core satisfies $\mathcal{C}^{(c)}\in\mathbb{R}^{r\times\cdots\times r}$.
Each TT core slice has dimensions
\begin{equation}
A_a^{(c)}(\nu_a)\in\mathbb{R}^{r_{a-1}\times r_a},\qquad
r_0=r_{n_{\mathrm{ax}}}=1,\qquad
r_a=r\quad(1\leq a<n_{\mathrm{ax}}).
\end{equation}

Here, $\otimes$ and $\times_a$ denote the outer product and the mode-$a$ tensor--matrix product, respectively. CP represents a tensor as a sum of rank-one tensors, whereas Tucker combines a core tensor with axis-specific factor matrices~\citep{kolda2009tensor}. In this work, CP uses $r$ components, and Tucker uses rank $r$ along each axis with a core containing $r^{n_{\mathrm{ax}}}$ entries. TT represents a tensor as a sequential product of cores~\citep{oseledets2011tensor}; we set both boundary ranks to 1 and all internal ranks to $r$. The CP and Tucker generators output $r$ values per coordinate, while TT outputs $r$ values at the two boundary axes and $r^2$ values at internal axes. These ranks specify the decompositions of the individual real-valued tensors and do not denote the exact rank of the final complex-valued kernel.

An independent CAFE+ factor generator is used for each axis and each real or imaginary component. For CP and Tucker, the midpoint coordinate $s_{I_a}(\nu_a)$ from Appendix~\ref{app:coordinates} is supplied as input, and the generator outputs $r$ values corresponding to one row of the factor matrix. For TT, the generator similarly outputs $r_{a-1}r_a$ values, which are reshaped into a core slice. The reconstructed tensor, whose axes are ordered as input components, frequency-storage axes, and output components, is rearranged into an output--input matrix for each mode and used for spectral mixing.

\section{Model Settings}
\label{app:settings}

CAFE+FNO uses a hidden channel width of 32, four operator layers, and two linear branches. The dense model uses \(m=32\) Fourier frequency vectors in \(\mathbb{R}^{d_s}\) and \(n=32\) Chebyshev polynomials along each coordinate axis, with a kernel MLP containing one hidden layer. The dataset-specific branch output dimensions, kernel MLP hidden widths, and decomposition ranks are listed in Table~\ref{tab:settings}.

Each factor generator in the decomposed models uses 16 Fourier frequency vectors and eight Chebyshev polynomials, with a branch output dimension of 12. The factor MLP hidden widths are 16, 24, and 16 for CP, TT, and Tucker, respectively. For the factorized SirenFNO baselines, the backbone input
dimension is 16 for all ReacDiff variants and for Tucker on NS,
and 32 otherwise.

\begin{table}[!htbp]
\caption{Dataset-specific architectural settings for CAFE+FNO. The branch output dimension and MLP hidden width refer to the dense model; the common settings for the decomposed models are described in the text.}
\label{tab:settings}
\begin{center}
\begin{tabularx}{\textwidth}{@{}l*{5}{>{\centering\arraybackslash}X}@{}}
\toprule
\textbf{Dataset} & \textbf{Dense branch output dimension} & \textbf{Dense MLP hidden width} & \textbf{CP rank} & \textbf{TT rank} & \textbf{Tucker per-axis rank} \\
\midrule
Darcy & 32 & 32 & 8 & 8 & 10 \\
Navier--Stokes & 32 & 64 & 16 & 16 & 16 \\
Burgers & 32 & 32 & 8 & 8 & 8 \\
Airfoil & 32 & 32 & 8 & 8 & 8 \\
ReacDiff & 16 & 32 & 8 & 8 & 8 \\
\bottomrule
\end{tabularx}
\end{center}
\end{table}

\begin{table}[!htbp]
\caption{Trainable parameter counts in real scalars, including
lifting, operator layers, and projection. Each complex parameter
counts as two real scalars; shared parameters are counted once.
Fixed buffers are excluded, and kernel-generator outputs are not
counted as additional parameters. The dash denotes an omitted
model--dataset combination.}
\label{tab:parameters}
\begin{center}
\begin{tabularx}{\textwidth}{@{}>{\raggedright\arraybackslash}X*{5}{c}@{}}
\toprule
\textbf{Model} & \textbf{Darcy} & \textbf{NS} & \textbf{Burgers} & \textbf{Airfoil} & \textbf{ReacDiff} \\
\midrule
\textbf{Resolution} & $128\times128$ & $128\times128$ & $1024$ & $221\times51$ & $1024$ \\
\midrule
FNO
& 1,192,801 & 4,469,601 & 4,216,161 & 2,372,513 & 4,216,161 \\

U-FNO
& 5,931,201 & 990,753 & 1,883,073 & --- & 1,892,161 \\

TFNO-CP
& 72,913 & 237,505 & 226,369 & 131,993 & 226,369 \\

AM-FNO (MLP)
& 385,473 & 4,443,137 & 823,073 & 1,136,673 & 823,073 \\

SirenFNO
& 308,865 & 579,457 & 308,993 & 308,897 & 304,833 \\

\textbf{CAFE+FNO}
& 345,601 & 611,969 & 337,665 & 345,633 & 323,201 \\

CP-SirenFNO
& 80,513 & 138,881 & 70,145 & 80,545 & 57,665 \\

TT-SirenFNO
& 109,185 & 211,585 & 84,481 & 109,217 & 72,001 \\

Tucker-SirenFNO
& 162,561 & 596,353 & 74,241 & 113,313 & 61,761 \\

\textbf{CAFE+FNO (CP)}
& 80,517 & 84,869 & 70,149 & 80,549 & 70,149 \\

\textbf{CAFE+FNO (TT)}
& 108,293 & 188,293 & 85,381 & 108,325 & 85,381 \\

\textbf{CAFE+FNO (Tucker)}
& 161,605 & 609,157 & 74,245 & 113,317 & 74,245 \\
\bottomrule
\end{tabularx}
\end{center}
\end{table}

\section{Per-Seed Test Errors}
\label{app:per-seed-results}

Tables~\ref{tab:results-seed-0}--\ref{tab:results-seed-202}
provide the individual-seed test relative $L_2$ errors for the
models and datasets compared in Table~\ref{tab:results}, using
random seeds $\{0,42,73,108,202\}$. Each result is obtained from
the final-epoch model after 500 training epochs, following the
same evaluation protocols as the main comparison. Errors are
reported in scientific notation with three decimal places
in the mantissa.

\begin{table}[!htbp]
\caption{Test relative $L_2$ errors for seed $0$ after 500 training epochs. The dash denotes an omitted model--dataset combination.}
\label{tab:results-seed-0}
\begin{center}
\begin{tabular*}{\textwidth}{@{\extracolsep{\fill}}lccccc@{}}
\toprule
Model & Darcy & NS & Burgers & Airfoil & ReacDiff \\
\midrule
Resolution & $128\!\times\!128$ & $128\!\times\!128$ & $1024$ & $221\!\times\!51$ & $1024$ \\
\midrule
FNO & $6.200\mathrm{e}{-2}$ & $5.610\mathrm{e}{-2}$ & $1.135\mathrm{e}{-2}$ & $6.047\mathrm{e}{-3}$ & $4.252\mathrm{e}{-3}$ \\
U-FNO & $4.790\mathrm{e}{-2}$ & $5.824\mathrm{e}{-2}$ & $1.005\mathrm{e}{-2}$ & --- & $7.578\mathrm{e}{-3}$ \\
TFNO-CP & $4.928\mathrm{e}{-2}$ & $3.394\mathrm{e}{-2}$ & $9.423\mathrm{e}{-3}$ & $7.026\mathrm{e}{-3}$ & $4.185\mathrm{e}{-3}$ \\
AM-FNO (MLP) & $7.497\mathrm{e}{-2}$ & $3.113\mathrm{e}{-2}$ & $2.338\mathrm{e}{-2}$ & $6.519\mathrm{e}{-3}$ & $1.001\mathrm{e}{-2}$ \\
SirenFNO & $2.587\mathrm{e}{-2}$ & $4.460\mathrm{e}{-2}$ & $5.392\mathrm{e}{-3}$ & $5.691\mathrm{e}{-3}$ & $2.148\mathrm{e}{-3}$ \\
CAFE+FNO & $1.384\mathrm{e}{-2}$ & $1.756\mathrm{e}{-2}$ & $4.205\mathrm{e}{-3}$ & $4.707\mathrm{e}{-3}$ & $3.843\mathrm{e}{-3}$ \\
\midrule
CP-SirenFNO & $3.091\mathrm{e}{-2}$ & $4.714\mathrm{e}{-2}$ & $8.479\mathrm{e}{-3}$ & $6.910\mathrm{e}{-3}$ & $2.880\mathrm{e}{-3}$ \\
TT-SirenFNO & $1.428\mathrm{e}{-1}$ & $3.924\mathrm{e}{-2}$ & $7.403\mathrm{e}{-3}$ & $5.902\mathrm{e}{-3}$ & $2.531\mathrm{e}{-3}$ \\
Tucker-SirenFNO & $3.744\mathrm{e}{-2}$ & $4.890\mathrm{e}{-2}$ & $8.293\mathrm{e}{-3}$ & $7.876\mathrm{e}{-3}$ & $2.992\mathrm{e}{-3}$ \\
CAFE+FNO (CP) & $3.025\mathrm{e}{-2}$ & $9.429\mathrm{e}{-2}$ & $8.656\mathrm{e}{-3}$ & $8.307\mathrm{e}{-3}$ & $4.243\mathrm{e}{-3}$ \\
CAFE+FNO (TT) & $2.652\mathrm{e}{-2}$ & $6.126\mathrm{e}{-2}$ & $8.234\mathrm{e}{-3}$ & $7.194\mathrm{e}{-3}$ & $3.988\mathrm{e}{-3}$ \\
CAFE+FNO (Tucker) & $2.781\mathrm{e}{-2}$ & $4.893\mathrm{e}{-2}$ & $8.911\mathrm{e}{-3}$ & $6.724\mathrm{e}{-3}$ & $4.077\mathrm{e}{-3}$ \\
\bottomrule
\end{tabular*}
\end{center}
\end{table}

\begin{table}[!htbp]
\caption{Test relative $L_2$ errors for seed $42$ after 500 training epochs. The dash denotes an omitted model--dataset combination.}
\label{tab:results-seed-42}
\begin{center}
\begin{tabular*}{\textwidth}{@{\extracolsep{\fill}}lccccc@{}}
\toprule
Model & Darcy & NS & Burgers & Airfoil & ReacDiff \\
\midrule
Resolution & $128\!\times\!128$ & $128\!\times\!128$ & $1024$ & $221\!\times\!51$ & $1024$ \\
\midrule
FNO & $7.201\mathrm{e}{-2}$ & $5.585\mathrm{e}{-2}$ & $1.121\mathrm{e}{-2}$ & $6.130\mathrm{e}{-3}$ & $4.125\mathrm{e}{-3}$ \\
U-FNO & $4.904\mathrm{e}{-2}$ & $5.822\mathrm{e}{-2}$ & $9.668\mathrm{e}{-3}$ & --- & $4.914\mathrm{e}{-3}$ \\
TFNO-CP & $4.734\mathrm{e}{-2}$ & $3.525\mathrm{e}{-2}$ & $1.054\mathrm{e}{-2}$ & $6.670\mathrm{e}{-3}$ & $3.930\mathrm{e}{-3}$ \\
AM-FNO (MLP) & $4.344\mathrm{e}{-2}$ & $2.858\mathrm{e}{-2}$ & $2.010\mathrm{e}{-2}$ & $5.792\mathrm{e}{-3}$ & $1.052\mathrm{e}{-1}$ \\
SirenFNO & $2.285\mathrm{e}{-2}$ & $4.578\mathrm{e}{-2}$ & $5.617\mathrm{e}{-3}$ & $6.068\mathrm{e}{-3}$ & $2.384\mathrm{e}{-3}$ \\
CAFE+FNO & $1.382\mathrm{e}{-2}$ & $1.695\mathrm{e}{-2}$ & $3.718\mathrm{e}{-3}$ & $4.582\mathrm{e}{-3}$ & $3.763\mathrm{e}{-3}$ \\
\midrule
CP-SirenFNO & $3.122\mathrm{e}{-2}$ & $5.328\mathrm{e}{-2}$ & $8.780\mathrm{e}{-3}$ & $9.428\mathrm{e}{-3}$ & $7.126\mathrm{e}{-3}$ \\
TT-SirenFNO & $2.904\mathrm{e}{-2}$ & $4.263\mathrm{e}{-2}$ & $7.820\mathrm{e}{-3}$ & $6.409\mathrm{e}{-3}$ & $2.697\mathrm{e}{-3}$ \\
Tucker-SirenFNO & $5.288\mathrm{e}{-2}$ & $3.904\mathrm{e}{-2}$ & $8.214\mathrm{e}{-3}$ & $6.280\mathrm{e}{-3}$ & $2.784\mathrm{e}{-3}$ \\
CAFE+FNO (CP) & $3.373\mathrm{e}{-2}$ & $4.288\mathrm{e}{-2}$ & $8.076\mathrm{e}{-3}$ & $7.653\mathrm{e}{-3}$ & $4.391\mathrm{e}{-3}$ \\
CAFE+FNO (TT) & $2.777\mathrm{e}{-2}$ & $3.364\mathrm{e}{-2}$ & $8.554\mathrm{e}{-3}$ & $6.576\mathrm{e}{-3}$ & $4.449\mathrm{e}{-3}$ \\
CAFE+FNO (Tucker) & $3.034\mathrm{e}{-2}$ & $3.540\mathrm{e}{-2}$ & $8.844\mathrm{e}{-3}$ & $7.880\mathrm{e}{-3}$ & $4.367\mathrm{e}{-3}$ \\
\bottomrule
\end{tabular*}
\end{center}
\end{table}

\begin{table}[!htbp]
\caption{Test relative $L_2$ errors for seed $73$ after 500 training epochs. The dash denotes an omitted model--dataset combination.}
\label{tab:results-seed-73}
\begin{center}
\begin{tabular*}{\textwidth}{@{\extracolsep{\fill}}lccccc@{}}
\toprule
Model & Darcy & NS & Burgers & Airfoil & ReacDiff \\
\midrule
Resolution & $128\!\times\!128$ & $128\!\times\!128$ & $1024$ & $221\!\times\!51$ & $1024$ \\
\midrule
FNO & $6.389\mathrm{e}{-2}$ & $5.573\mathrm{e}{-2}$ & $1.134\mathrm{e}{-2}$ & $6.310\mathrm{e}{-3}$ & $3.944\mathrm{e}{-3}$ \\
U-FNO & $4.521\mathrm{e}{-2}$ & $5.775\mathrm{e}{-2}$ & $1.011\mathrm{e}{-2}$ & --- & $1.927\mathrm{e}{-2}$ \\
TFNO-CP & $4.770\mathrm{e}{-2}$ & $3.488\mathrm{e}{-2}$ & $1.030\mathrm{e}{-2}$ & $6.828\mathrm{e}{-3}$ & $3.715\mathrm{e}{-3}$ \\
AM-FNO (MLP) & $6.956\mathrm{e}{-2}$ & $2.806\mathrm{e}{-2}$ & $1.490\mathrm{e}{-2}$ & $7.247\mathrm{e}{-3}$ & $8.633\mathrm{e}{0}$ \\
SirenFNO & $2.517\mathrm{e}{-2}$ & $4.055\mathrm{e}{-2}$ & $6.347\mathrm{e}{-3}$ & $5.532\mathrm{e}{-3}$ & $3.682\mathrm{e}{-3}$ \\
CAFE+FNO & $1.362\mathrm{e}{-2}$ & $1.694\mathrm{e}{-2}$ & $4.514\mathrm{e}{-3}$ & $4.785\mathrm{e}{-3}$ & $4.203\mathrm{e}{-3}$ \\
\midrule
CP-SirenFNO & $4.091\mathrm{e}{-2}$ & $5.295\mathrm{e}{-2}$ & $9.258\mathrm{e}{-3}$ & $7.934\mathrm{e}{-3}$ & $3.116\mathrm{e}{-3}$ \\
TT-SirenFNO & $4.308\mathrm{e}{-2}$ & $3.731\mathrm{e}{-2}$ & $7.753\mathrm{e}{-3}$ & $5.915\mathrm{e}{-3}$ & $3.006\mathrm{e}{-3}$ \\
Tucker-SirenFNO & $7.154\mathrm{e}{-2}$ & $4.141\mathrm{e}{-2}$ & $8.643\mathrm{e}{-3}$ & $1.482\mathrm{e}{-2}$ & $3.007\mathrm{e}{-3}$ \\
CAFE+FNO (CP) & $2.715\mathrm{e}{-2}$ & $5.776\mathrm{e}{-2}$ & $8.819\mathrm{e}{-3}$ & $7.006\mathrm{e}{-3}$ & $3.579\mathrm{e}{-3}$ \\
CAFE+FNO (TT) & $2.944\mathrm{e}{-2}$ & $5.075\mathrm{e}{-2}$ & $8.419\mathrm{e}{-3}$ & $8.021\mathrm{e}{-3}$ & $3.453\mathrm{e}{-3}$ \\
CAFE+FNO (Tucker) & $2.702\mathrm{e}{-2}$ & $3.660\mathrm{e}{-2}$ & $9.431\mathrm{e}{-3}$ & $9.674\mathrm{e}{-3}$ & $3.735\mathrm{e}{-3}$ \\
\bottomrule
\end{tabular*}
\end{center}
\end{table}

\begin{table}[!htbp]
\caption{Test relative $L_2$ errors for seed $108$ after 500 training epochs. The dash denotes an omitted model--dataset combination.}
\label{tab:results-seed-108}
\begin{center}
\begin{tabular*}{\textwidth}{@{\extracolsep{\fill}}lccccc@{}}
\toprule
Model & Darcy & NS & Burgers & Airfoil & ReacDiff \\
\midrule
Resolution & $128\!\times\!128$ & $128\!\times\!128$ & $1024$ & $221\!\times\!51$ & $1024$ \\
\midrule
FNO & $6.690\mathrm{e}{-2}$ & $5.584\mathrm{e}{-2}$ & $1.122\mathrm{e}{-2}$ & $6.000\mathrm{e}{-3}$ & $4.239\mathrm{e}{-3}$ \\
U-FNO & $4.812\mathrm{e}{-2}$ & $5.636\mathrm{e}{-2}$ & $1.008\mathrm{e}{-2}$ & --- & $7.520\mathrm{e}{-3}$ \\
TFNO-CP & $4.607\mathrm{e}{-2}$ & $3.504\mathrm{e}{-2}$ & $1.046\mathrm{e}{-2}$ & $7.891\mathrm{e}{-3}$ & $4.407\mathrm{e}{-3}$ \\
AM-FNO (MLP) & $5.421\mathrm{e}{-2}$ & $3.245\mathrm{e}{-2}$ & $1.719\mathrm{e}{-2}$ & $5.511\mathrm{e}{-3}$ & $1.133\mathrm{e}{-2}$ \\
SirenFNO & $2.442\mathrm{e}{-2}$ & $4.266\mathrm{e}{-2}$ & $5.710\mathrm{e}{-3}$ & $5.936\mathrm{e}{-3}$ & $1.789\mathrm{e}{-3}$ \\
CAFE+FNO & $1.353\mathrm{e}{-2}$ & $1.753\mathrm{e}{-2}$ & $3.760\mathrm{e}{-3}$ & $4.588\mathrm{e}{-3}$ & $3.452\mathrm{e}{-3}$ \\
\midrule
CP-SirenFNO & $4.386\mathrm{e}{-2}$ & $4.228\mathrm{e}{-2}$ & $9.003\mathrm{e}{-3}$ & $6.144\mathrm{e}{-3}$ & $4.083\mathrm{e}{-3}$ \\
TT-SirenFNO & $6.327\mathrm{e}{-1}$ & $3.225\mathrm{e}{-2}$ & $7.656\mathrm{e}{-3}$ & $5.744\mathrm{e}{-3}$ & $2.533\mathrm{e}{-3}$ \\
Tucker-SirenFNO & $1.073\mathrm{e}{-1}$ & $6.707\mathrm{e}{-2}$ & $7.504\mathrm{e}{-3}$ & $5.547\mathrm{e}{-3}$ & $2.928\mathrm{e}{-3}$ \\
CAFE+FNO (CP) & $3.098\mathrm{e}{-2}$ & $1.011\mathrm{e}{-1}$ & $8.024\mathrm{e}{-3}$ & $1.312\mathrm{e}{-2}$ & $4.767\mathrm{e}{-3}$ \\
CAFE+FNO (TT) & $2.546\mathrm{e}{-2}$ & $3.159\mathrm{e}{-2}$ & $7.997\mathrm{e}{-3}$ & $7.036\mathrm{e}{-3}$ & $4.481\mathrm{e}{-3}$ \\
CAFE+FNO (Tucker) & $2.979\mathrm{e}{-2}$ & $3.501\mathrm{e}{-2}$ & $9.368\mathrm{e}{-3}$ & $7.324\mathrm{e}{-3}$ & $3.454\mathrm{e}{-3}$ \\
\bottomrule
\end{tabular*}
\end{center}
\end{table}

\begin{table}[!htbp]
\caption{Test relative $L_2$ errors for seed $202$ after 500 training epochs. The dash denotes an omitted model--dataset combination.}
\label{tab:results-seed-202}
\begin{center}
\begin{tabular*}{\textwidth}{@{\extracolsep{\fill}}lccccc@{}}
\toprule
Model & Darcy & NS & Burgers & Airfoil & ReacDiff \\
\midrule
Resolution & $128\!\times\!128$ & $128\!\times\!128$ & $1024$ & $221\!\times\!51$ & $1024$ \\
\midrule
FNO & $6.220\mathrm{e}{-2}$ & $5.794\mathrm{e}{-2}$ & $1.138\mathrm{e}{-2}$ & $6.387\mathrm{e}{-3}$ & $4.347\mathrm{e}{-3}$ \\
U-FNO & $4.499\mathrm{e}{-2}$ & $5.789\mathrm{e}{-2}$ & $1.020\mathrm{e}{-2}$ & --- & $6.383\mathrm{e}{-3}$ \\
TFNO-CP & $4.631\mathrm{e}{-2}$ & $3.660\mathrm{e}{-2}$ & $1.064\mathrm{e}{-2}$ & $6.896\mathrm{e}{-3}$ & $4.123\mathrm{e}{-3}$ \\
AM-FNO (MLP) & $6.792\mathrm{e}{-2}$ & $2.946\mathrm{e}{-2}$ & $1.809\mathrm{e}{-2}$ & $6.375\mathrm{e}{-3}$ & $3.048\mathrm{e}{-2}$ \\
SirenFNO & $2.341\mathrm{e}{-2}$ & $4.262\mathrm{e}{-2}$ & $6.246\mathrm{e}{-3}$ & $5.552\mathrm{e}{-3}$ & $1.774\mathrm{e}{-3}$ \\
CAFE+FNO & $1.370\mathrm{e}{-2}$ & $1.701\mathrm{e}{-2}$ & $4.440\mathrm{e}{-3}$ & $4.856\mathrm{e}{-3}$ & $3.731\mathrm{e}{-3}$ \\
\midrule
CP-SirenFNO & $8.353\mathrm{e}{-2}$ & $6.165\mathrm{e}{-2}$ & $9.014\mathrm{e}{-3}$ & $1.018\mathrm{e}{-2}$ & $4.658\mathrm{e}{-3}$ \\
TT-SirenFNO & $4.625\mathrm{e}{-2}$ & $2.878\mathrm{e}{-2}$ & $7.972\mathrm{e}{-3}$ & $6.147\mathrm{e}{-3}$ & $2.401\mathrm{e}{-3}$ \\
Tucker-SirenFNO & $5.115\mathrm{e}{-2}$ & $3.709\mathrm{e}{-2}$ & $7.878\mathrm{e}{-3}$ & $6.363\mathrm{e}{-3}$ & $2.904\mathrm{e}{-3}$ \\
CAFE+FNO (CP) & $2.814\mathrm{e}{-2}$ & $4.846\mathrm{e}{-2}$ & $1.010\mathrm{e}{-2}$ & $7.332\mathrm{e}{-3}$ & $3.755\mathrm{e}{-3}$ \\
CAFE+FNO (TT) & $2.965\mathrm{e}{-2}$ & $3.525\mathrm{e}{-2}$ & $9.191\mathrm{e}{-3}$ & $6.499\mathrm{e}{-3}$ & $4.065\mathrm{e}{-3}$ \\
CAFE+FNO (Tucker) & $2.481\mathrm{e}{-2}$ & $4.197\mathrm{e}{-2}$ & $8.275\mathrm{e}{-3}$ & $9.050\mathrm{e}{-3}$ & $2.981\mathrm{e}{-3}$ \\
\bottomrule
\end{tabular*}
\end{center}
\end{table}

\clearpage
\section{Qualitative Comparisons on Random Test Samples}
\label{app:qualitative_comparisons}

We qualitatively compare FNO, AM-FNO, SirenFNO, and CAFE+FNO
on Darcy flow, Navier--Stokes, and Airfoil.
For each dataset, three test samples were selected at random,
and all four models were evaluated on the same selected samples
using checkpoints trained with seed 0.
The results are organized into three sample sets,
each containing one test sample from each dataset.

In all figures, rows correspond to Darcy flow, Navier--Stokes,
and Airfoil, from top to bottom.
Prediction figures show the ground truth followed by predictions
from FNO, AM-FNO, SirenFNO, and CAFE+FNO, from left to right.
The corresponding error maps show the pointwise absolute error
$|\hat{u}-u|$ for the four models in the same order,
where $\hat{u}$ and $u$ denote the prediction and ground truth.
For each figure, all panels within a dataset row share
the same color scale.
Values below the model panels report sample-wise relative $L_2$ errors.
Airfoil panels show a zoomed view around the airfoil.

\begin{figure}[htbp]
    \centering
    \includegraphics[width=\linewidth]
    {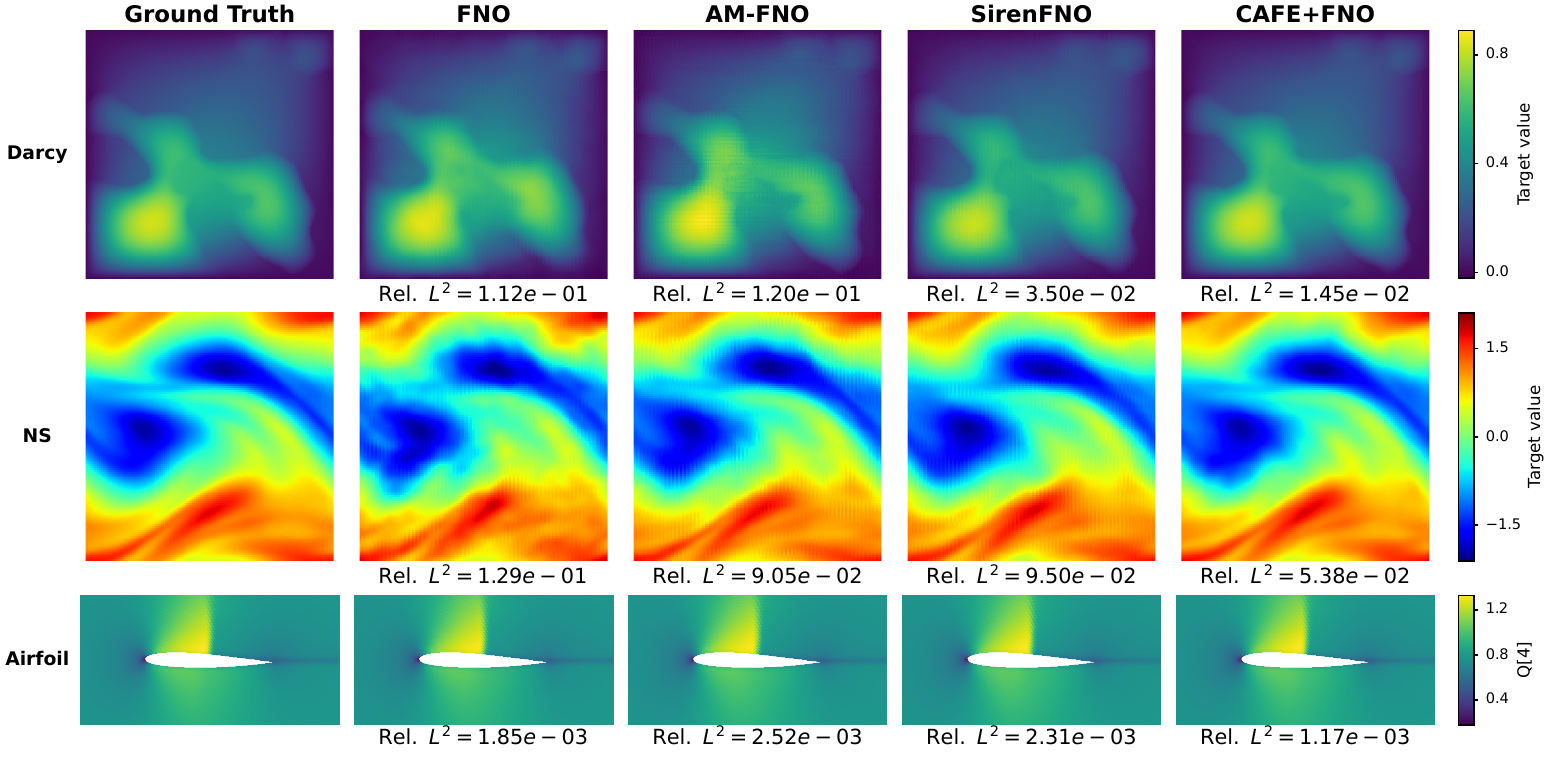}
    \caption{
    Ground truth and model predictions for random test sample set 1.
    }
    \label{fig:app_predictions_01}
\end{figure}
\begin{figure}[htbp]
    \centering
    \includegraphics[width=\linewidth]
    {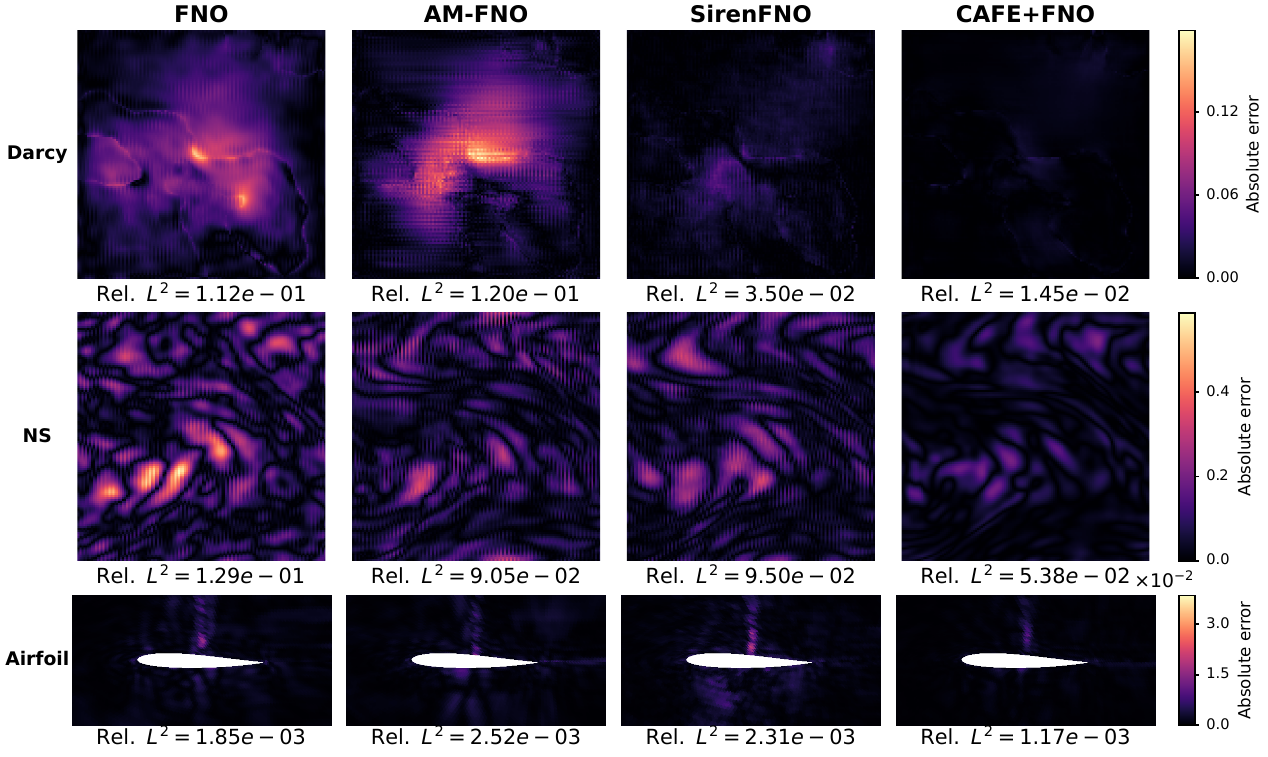}
    \caption{Absolute error maps for random test sample set 1.}
\label{fig:app_errors_01}
\end{figure}
\begin{figure}[htbp]
    \centering
    \includegraphics[width=\linewidth]
    {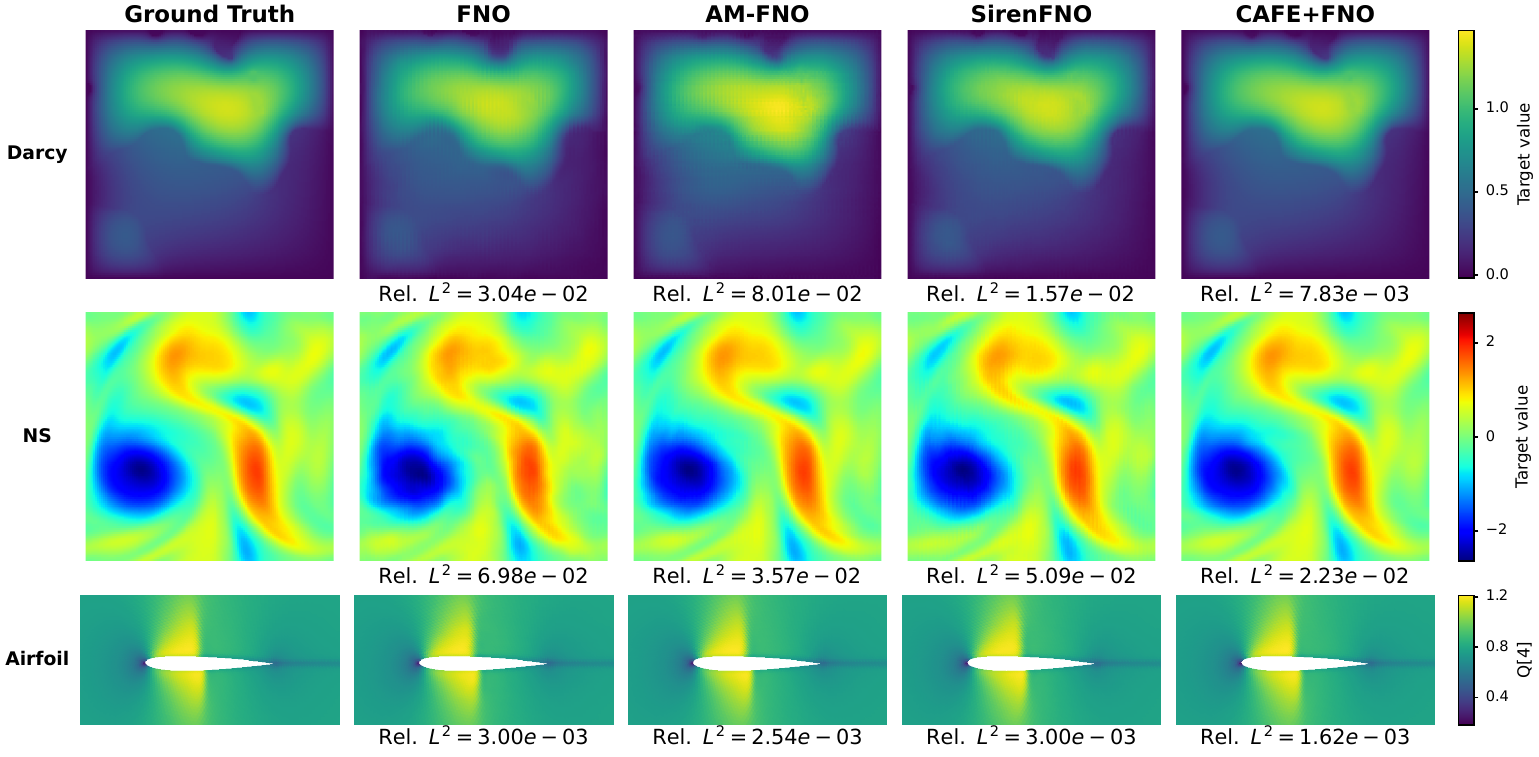}
    \caption{
    Ground truth and model predictions for random test sample set 2.
    }
    \label{fig:app_predictions_02}
\end{figure}
\begin{figure}[htbp]
    \centering
    \includegraphics[width=\linewidth]
    {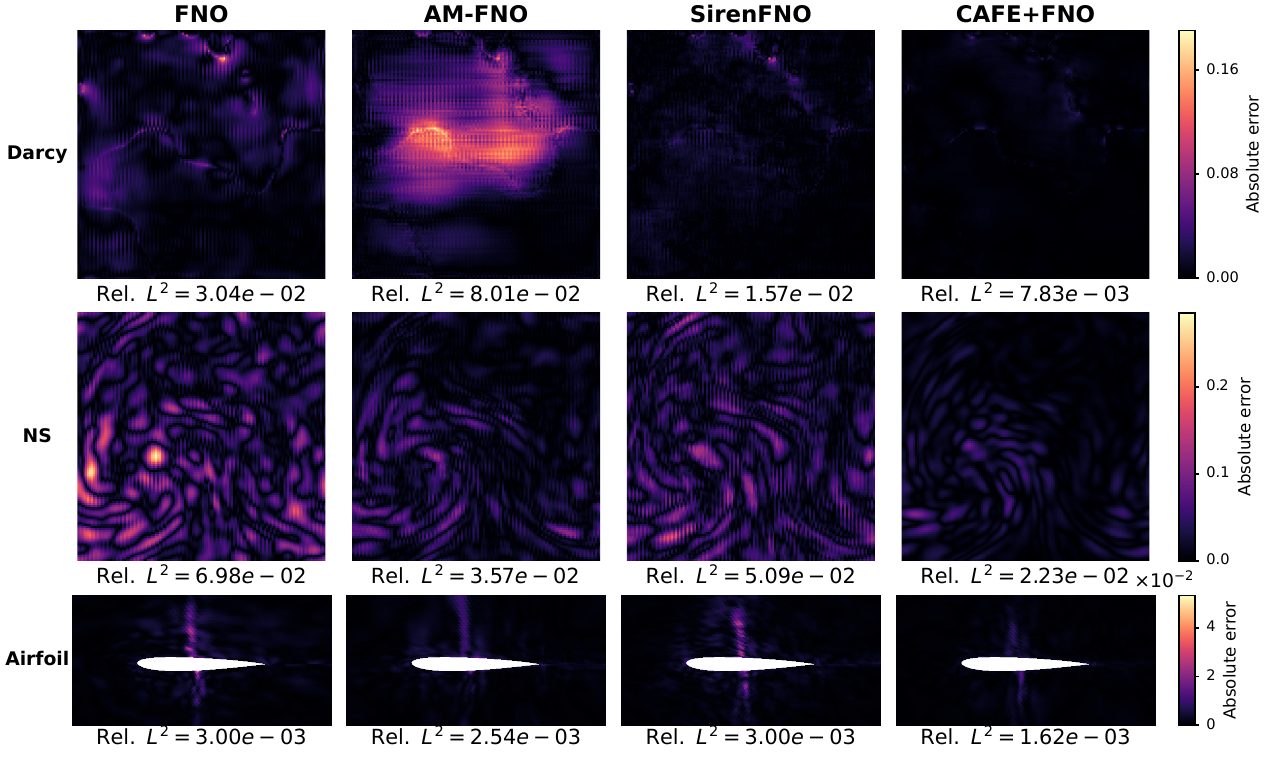}
    \caption{Absolute error maps for random test sample set 2.}
    \label{fig:app_errors_02}
\end{figure}
\begin{figure}[htbp]
    \centering
    \includegraphics[width=\linewidth]
    {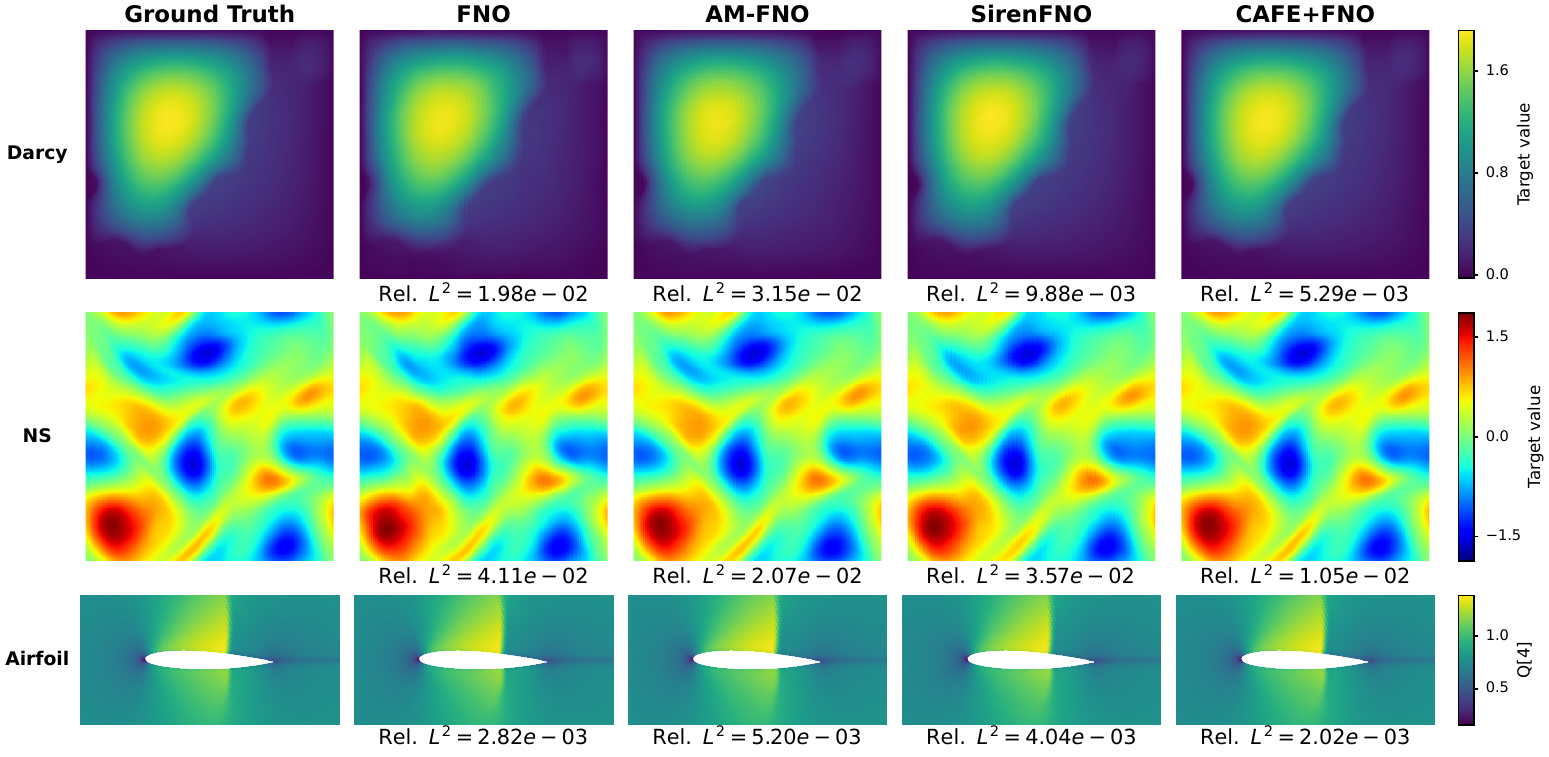}
    \caption{
    Ground truth and model predictions for random test sample set 3.
    }
    \label{fig:app_predictions_03}
\end{figure}
\begin{figure}[htbp]
    \centering
    \includegraphics[width=\linewidth]
    {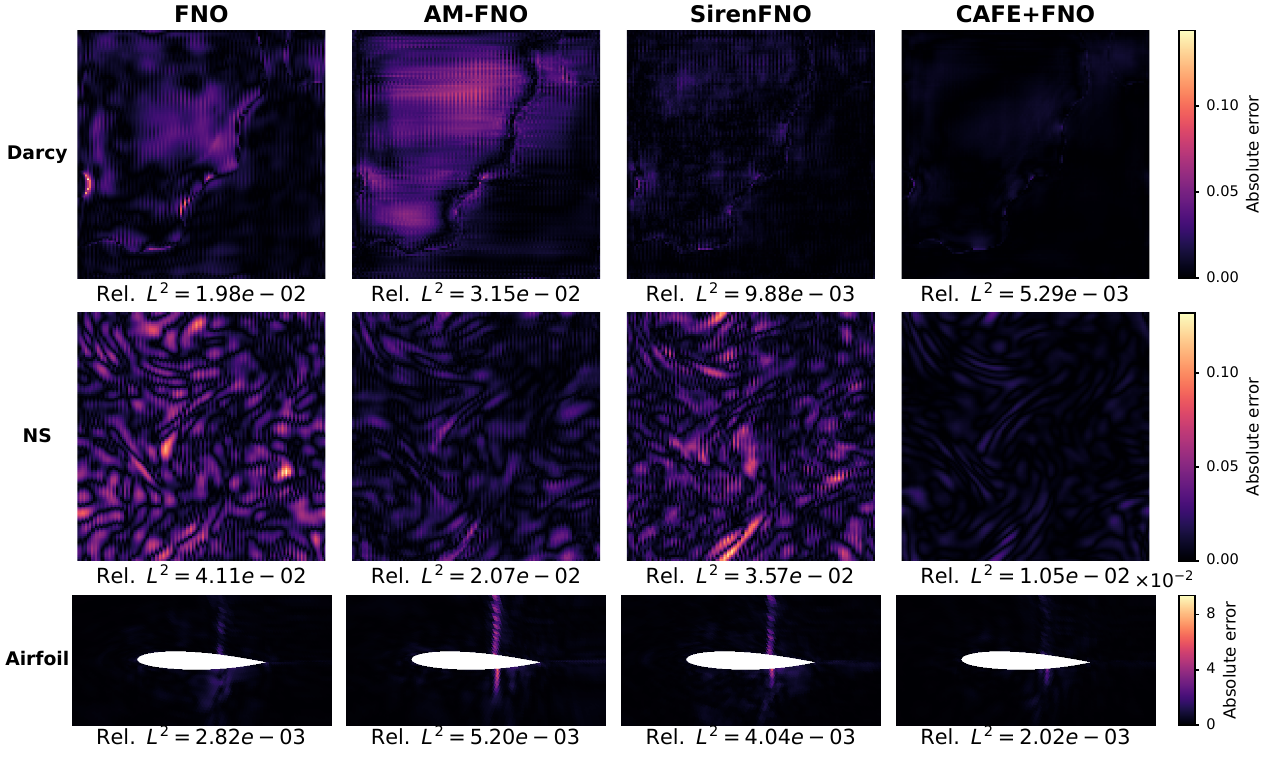}
    \caption{Absolute error maps for random test sample set 3.}
    \label{fig:app_errors_03}
\end{figure}

\clearpage
\end{document}